\documentclass[runningheads]{llncs}
\usepackage{eccv}

\usepackage{amsmath,amsfonts,bm}

\def\eqref#1{equation~\ref{#1}}

\def\1{\bm{1}}

\def\vc{{\bm{c}}}

\def\vn{{\bm{n}}}

\def\vr{{\bm{r}}}

\def\vx{{\bm{x}}}
\def\vy{{\bm{y}}}
\def\vz{{\bm{z}}}

\DeclareMathAlphabet{\mathsfit}{\encodingdefault}{\sfdefault}{m}{sl}
\SetMathAlphabet{\mathsfit}{bold}{\encodingdefault}{\sfdefault}{bx}{n}

\def\gR{{\mathcal{R}}}

\usepackage{url}
\usepackage{graphicx}
\usepackage{booktabs}
\usepackage[table]{xcolor}
\usepackage{multirow}
\usepackage[ruled,vlined]{algorithm2e}
\usepackage{amssymb} 
\usepackage{mathtools}
\usepackage{minted}
\usepackage{wrapfig}
\usepackage{authblk}
\usepackage{tocloft}
\usepackage[pagebackref,breaklinks,colorlinks,citecolor=eccvblue]{hyperref}
\usepackage[accsupp]{axessibility}
\usepackage{orcidlink}

\begin{document}

\title{LatSearch: Latent Reward-Guided Search for Faster Inference-Time Scaling in Video Diffusion} 
\titlerunning{LatSearch}
\author{Zengqun~Zhao$^{1}$\orcidlink{0000-0002-7135-1946}, Ziquan~Liu$^{1}$\orcidlink{0000-0002-9724-2510}, Yu~Cao$^{1}$\orcidlink{0009-0003-9128-1775}, Shaogang~Gong$^{1}$\orcidlink{0000-0001-8156-2299}, Zhensong Zhang$^{3}$\orcidlink{0009-0001-7911-7564}, Jifei~Song$^{3}$\orcidlink{0000-0002-3381-6685}, Jiankang~Deng$^{2}$\orcidlink{0000-0002-3709-6216}, Ioannis~Patras$^{1}$\orcidlink{0000-0003-3913-4738}}
\institute{$^{1}$Queen Mary University of London, $^{2}$Imperial College London, $^{3}$Huawei R\&D UK \\
\url{https://zengqunzhao.github.io/LatSearch}}
\authorrunning{Z.~Zhao et al.}
\maketitle

\vspace{-1em}
\begin{figure}[h]
    \centering 
    \includegraphics[width=0.99\textwidth]{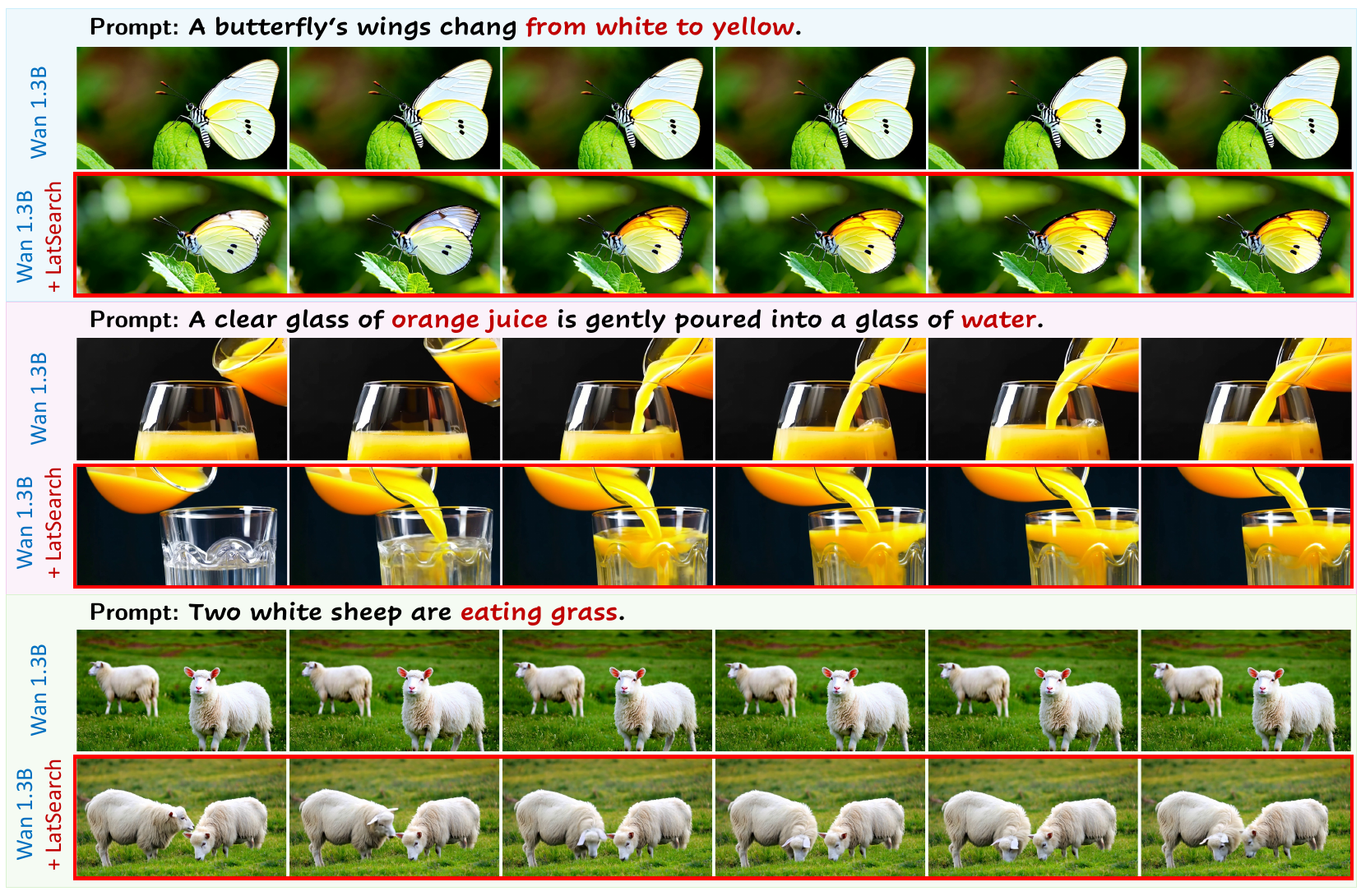}
    \vspace{-1em}
    \caption{Text-to-video generations, comparing a vanilla model with LatSearch, a novel faster inference-time scaling method in video generation. LatSearch significantly improves sample quality by leveraging latent reward-guided computation allocation during inference, enabling early evaluation of noisy latents and the selection of credible candidates along the diffusion trajectory.}
    \label{fig1}
\end{figure}
\vspace{-3em}

\begin{abstract}
The recent success of inference-time scaling in large language models has inspired similar explorations in video diffusion. In particular, motivated by the existence of “golden noise” that enhances video quality, prior work has attempted to improve inference by optimising or searching for better initial noise. However, these approaches have notable limitations: they either rely on priors imposed at the beginning of noise sampling or on rewards evaluated only on the denoised and decoded videos. This leads to error accumulation, delayed and sparse reward signals, and prohibitive computational cost, which prevents the use of stronger search algorithms. Crucially, stronger search algorithms are precisely what could unlock substantial gains in controllability, sample efficiency and generation quality for video diffusion, provided their computational cost can be reduced. To fill in this gap, we enable efficient inference-time scaling for video diffusion through latent reward guidance, which provides intermediate, informative and efficient feedback along the denoising trajectory. We introduce a latent reward model that scores partially denoised latents at arbitrary timesteps with respect to visual quality, motion quality, and text alignment. Building on this model, we propose LatSearch, a novel inference-time search mechanism that performs Reward-Guided Resampling and Pruning (RGRP). In the resampling stage, candidates are sampled according to reward-normalised probabilities to reduce over-reliance on the reward model. In the pruning stage, applied at the final scheduled step, only the candidate with the highest cumulative reward is retained, improving both quality and efficiency. We evaluate LatSearch on the VBench-2.0 benchmark and demonstrate that it consistently improves video generation across multiple evaluation dimensions compared to the baseline Wan2.1 model. Compared with the state-of-the-art, our approach achieves comparable or better quality while reducing runtime by up to 79\%. 
\keywords{Video generation \and Diffusion models \and Latent reward models}
\end{abstract}
\vspace{-1em}
\vspace{-1.0em}
\section{Introduction}
\vspace{-0.5em}
Given the wide range of applications of video generation, such as video editing~\cite{kara2024rave}, customisation~\cite{karras2023dreampose}, image animation~\cite{dalal2025one}, and world modelling~\cite{agarwal2025cosmos}, there has been a growing interest in transferring the success of inference-time scaling observed in large language models (LLMs) to diffusion-based video generation models~\cite{liu2025video,ma2025scaling}. A natural way for better fidelity is to increase the number of denoising steps, which directly improves sample fidelity. However, recent research has shown that inference-time scaling extends further than simply increasing denoising steps~\cite{ma2025scaling}. Several studies have demonstrated the effectiveness of so-called “golden noise”, specific initial noise realisations that reliably lead to higher-quality generations, highlighting the significant role of noise initialisation in the final generation quality~\cite{zhou2024golden,qi2024not,bancrystal,kim2025model}. Consequently, many works now attempt to allocate additional computation at inference time to improve the fidelity and consistency of generated videos~\cite{oshima2025inference,he2025scaling,yang2025scalingnoise,ma2025scaling}. 

A key problem in inference-time scaling for video diffusion lies in the inability to evaluate intermediate latents reliably. Lacking such evaluations prohibits a model from supporting more flexible strategies, such as early stopping for efficiency, resulting in errors introduced at the initial stage being accumulated throughout the long denoising trajectory. Existing methods, however, largely overlook this problem. Noise optimisation techniques bias the initialisation by injecting noise into a reference video, applying temporal warping or optical flow, or fusing frequency components of denoised latents~\cite{chang2024warped,wu2024freeinit,burgert2025go,yuan2025freqprior,zhang2025training}. However, once the trajectory begins, they lack mechanisms to monitor and correct intermediate states. Noise search methods instead generate multiple candidates and select the best based on fully decoded videos. Such models leverage strategies such as Best-of-N sampling, beam search, evolutionary algorithms, or path search~\cite{singhal2025general,oshima2025inference,yang2025scalingnoise,he2025scaling,ma2025scaling}. These methods typically depend on reward models or verifiers, which are chosen from standard metrics (FID, IS, DINO, CLIP), or video-specific reward functions~\cite{liu2025improving}. More recently, uncertainty measures derived from attention maps have also been considered~\cite{kim2025model}. However, current noise search methods operate only on final outputs, causing them to incur high computational cost from full video decoding and suffer from reward delay~\cite{liao2025step}, limiting their usefulness in guiding generation.

To address this limitation, we propose LatSearch, a faster and better inference time scaling method that integrates a latent reward model into video diffusion. Unlike conventional verifiers that operate only on final decoded videos, our reward model evaluates partially denoised latents at arbitrary timesteps, providing intermediate feedback on generation progress to facilitate efficient inference-time search. By introducing process-level supervision, LatSearch can identify and prune low-quality candidates early, thereby reducing unnecessary denoising steps. This not only mitigates error accumulation and reward delay but also avoids the heavy cost of repeatedly decoding full videos, making a more efficient and effective search procedure possible. To enable inference-time search guided by intermediate latents, we propose a latent reward model that evaluates partially denoised latents at arbitrary timesteps with respect to visual quality, motion quality, and text alignment. This model directly guides the search process by scoring intermediate latents, enabling more fine-grained candidate selection than final-video evaluation. For training, we construct a dataset of (prompt, latent, timestep, video score, latent similarity) tuples, where latent similarity measures the correspondence between an intermediate latent and the final clean latent. The model is optimised with both a regression loss and a latent preference loss to improve accuracy and robustness in intermediate reward estimation. Building on this model, LatSearch introduces Reward-Guided Resampling and Pruning (RGRP) to refine noise candidates during generation. The resampling stage is inspired by importance sampling: instead of deterministically picking the highest-reward candidate, we sample candidates according to reward-normalised probabilities. This prevents over-reliance on the reward model, which may be imperfect. The pruning stage, applied at the final scheduled timestep, selects the single candidate with the highest cumulative reward across timesteps, reducing redundancy and significantly lowering the computational cost of the remaining denoising process.

In summary, the main contributions of this work are as follows:
\vspace{-1em}
\begin{itemize}
\item We propose a reward model that evaluates partially denoised latents at arbitrary timesteps, providing process-level supervision on visual quality, motion quality, and text alignment. Since intermediate latents lack explicit semantics, we introduce a similarity-based grounding strategy and combine regression with preference losses to make latent-level reward estimation feasible.
\item Building on this reward model, we design an inference-time scaling algorithm that incorporates intermediate supervision directly into the denoising trajectory. Candidates are probabilistically resampled according to reward-normalised weights, duplicates are removed via uniqueness pruning, and cumulative rewards are used for final selection, making latent reward actionable for efficient search.
\item On VBench2.0, LatSearch consistently improves video generation across creativity, commonsense, controllability, human fidelity, and physics. Compared with state-of-the-art inference-time scaling methods, our approach achieves comparable or better quality while reducing runtime by up to 79\%.  
\end{itemize}
\vspace{-0.5em}
\vspace{-1em}
\section{Related Work}
\vspace{-0.5em}
We review related work in two areas most relevant to our study: video generation methods and inference-time scaling for diffusion models.

\noindent \textbf{Video Generation.} Early studies on video generation explored diverse parad-igms, including VAEs~\cite{bhagat2020disentangling,skorokhodov2022stylegan}, GANs~\cite{hsieh2018learning,brooks2022generating}, and autoregressive models \cite{deng2024autoregressive,gu2025long}. More recently, diffusion-based approaches have become the dominant paradigm \cite{guo2024animatediff,yangcogvideox,dalal2025one}, achieving superior visual fidelity and scalability by extending the success of image diffusion models. Progress in video diffusion has generally followed two directions: (i) foundational models, which adapt image diffusion architectures to the temporal domain, and (ii) enhancements, which improve fidelity, efficiency, and controllability. In the text-to-video (T2V) setting, early models extended U-Net–based image diffusion backbones with temporal modules, as in Stable Video Diffusion (SVD)~\cite{blattmann2023stable}. Subsequent works introduced improved training strategies for temporal coherence and motion quality, exemplified by ModelScope~\cite{wang2023modelscope}, VideoCrafter~\cite{chen2023videocrafter1}, and AnimateDiff~\cite{guo2024animatediff}. More recently, Diffusion Transformers (DiTs)~\cite{peebles2023scalable} have emerged as stronger backbones, offering improved scalability and modelling capacity. At the same time, training objectives have evolved beyond conventional denoising losses toward flow-based formulations, enabling more efficient and stable optimisation. Works such as UViT~\cite{bao2023all} and Gentron \cite{chen2024gentron} first demonstrated the feasibility of Transformer-only backbones, inspiring large-scale systems like Hunyuan Video~\cite{kong2024hunyuanvideo}, CogVideoX~\cite{yangcogvideox}, and Wan~\cite{wan2025wan}. Building on these foundation models, subsequent research has explored generating longer videos~\cite{qiu2023freenoise,ma2025tuning,ouyang2025tokensgen}, improving temporal coherence~\cite{luo2025enhance,peruzzo2025ragme,nam2025optical}, leveraging human feedback~\cite{yuan2024instructvideo,hiranakahero,liu2025improving,zhu2025aligning}, enhancing controllability~\cite{chen2023control,xiao2024video}, and improving generation efficiency~\cite{sun2025swiftvideo}. In our work, we build on the state-of-the-art Wan model as the foundation for video generation, and investigate inference-time scaling through latent reward guidance. While prior work has explored latent reward models for few-shot video generation~\cite{ding2024dollar}, they are limited to evaluating the final denoised latent. In contrast, our approach evaluates intermediate states across denoising timesteps, enabling more fine-grained and effective guidance.

\noindent \textbf{Inference-Time Scaling for Diffusion Models.} There is growing interest in inference-time scaling for diffusion models, inspired by its success in large language models (LLMs)~\cite{liu2025video,ma2025scaling}. A straightforward scaling strategy is to increase the number of denoising steps, which generally improves sample fidelity. However, recent studies show that inference-time scaling extends far beyond this~\cite{ma2025scaling}. In particular, the choice of initial noise has been identified as a key factor in generation quality, with the notion of “golden noise” underscoring its importance~\cite{zhou2024golden,qi2024not,bancrystal,kim2025model}. As a result, many methods now dedicate additional computation at inference time to either optimise the initial noise or search for better noise samples~\cite{oshima2025inference,he2025scaling,yang2025scalingnoise,ma2025scaling}. Noise optimisation approaches inject priors into noise initialisation, for example by adding noise to a reference video~\cite{zhang2025training}, warping noise via temporal correlation or optical flow~\cite{chang2024warped,burgert2025go}, or fusing frequency components of denoised latents~\cite{wu2024freeinit,yuan2025freqprior}. While fusion-based methods avoid reliance on external inputs, they often require more iterations. Noise search approaches instead generate multiple candidates and evaluate the resulting videos. Techniques include Best-of-N sampling~\cite{singhal2025general}, beam search~\cite{oshima2025inference,yang2025scalingnoise}, evolutionary search~\cite{he2025scaling}, and search-over-path strategies~\cite{ma2025scaling}. Candidate evaluation typically relies on reward models or verifiers, ranging from traditional metrics such as FID~\cite{heusel2017gans}, IS~\cite{salimans2016improved}, DINO~\cite{caron2021emerging}, and CLIP~\cite{radford2021learning}, to video-specific reward models designed for temporal quality assessment~\cite{liu2025improving}. Beyond explicit search, recent work has also explored estimating noise quality directly from model attention maps using uncertainty-based measures~\cite{kim2025model}. Existing inference-time scaling methods optimise or search over initial noise and evaluate only final videos, lacking intermediate guidance. In contrast, we propose a latent reward model that assesses partially denoised latents at arbitrary timesteps, providing fine-grained feedback on visual quality, motion, and text alignment.

\begin{figure}[t]
    \centering 
    \includegraphics[width=\textwidth]{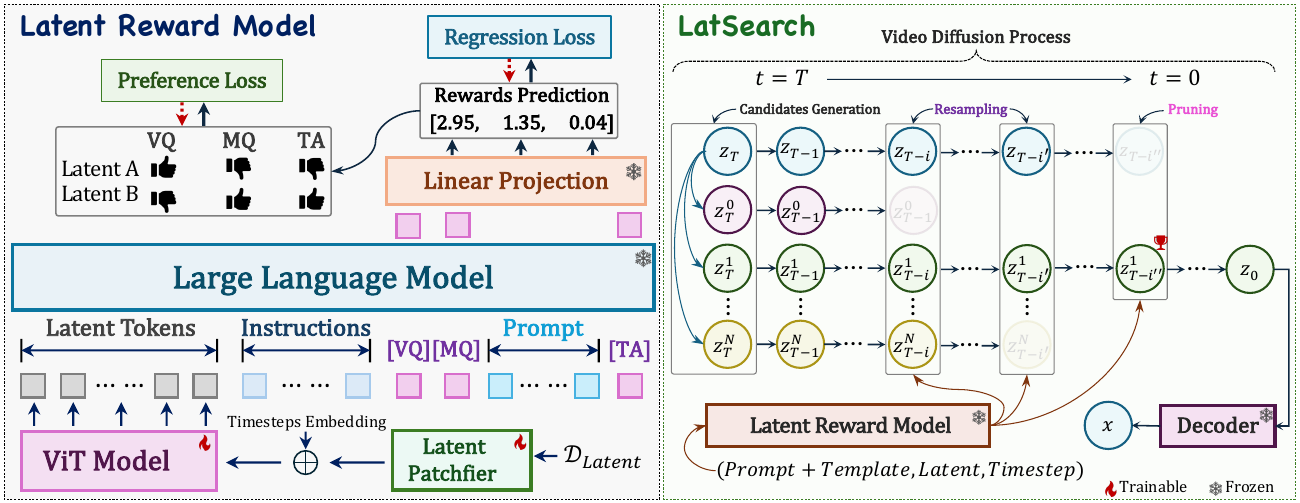}
    \vspace{-2em}
    \caption{An overview of a latent reward model (left) and the proposed latent reward-guided inference-time search method, LatSearch (right). On the left, input latent tokens are patchified, fused with timestep embeddings, and projected by a ViT encoder. Together with instruction tokens, text prompts, and special query tokens ([VQ], [MQ], [TA]), these form the input to a large language model. The model is trained using a combination of regression and preference losses. On the right, LatSearch maintains multiple candidate trajectories during a diffusion process. Candidates are periodically scored by the latent reward model, resampled with uniqueness to encourage diversity, and finally pruned 
based on cumulative rewards before decoding into the final video.} 
    \label{fig2}
    \vspace{-1.8em}
\end{figure}
\vspace{-1.0em}
\section{Method}
\vspace{-0.5em}
Our approach, LatSearch, integrates a latent reward model with a reward-guided search mechanism to scale video diffusion at inference time. In this section, we first review the preliminaries of video diffusion models, then introduce a latent reward model that provides intermediate evaluations throughout the denoising process, and finally describe how these signals are incorporated into a reward-guided search strategy for efficient and high-quality video generation.

\vspace{-1.0em}
\subsection{Preliminaries}
\vspace{-0.5em}
Latent text-to-video diffusion models encode an $F$-frame clean video $\{\vx^{(i)}\}_{i=1}^N \in \mathbb{R}^{F \times C \times H \times W}$ into latent representations $\{\vz^{(i)}_0\}_{i=1}^N$ using an encoder $\mathcal{E}$, where $C, H, W$ denote the channel, height, and width of each frame. For convenience, we denote
$\vz_0=\{\vz^{(i)}_0\}_{i=1}^N \sim p_0(\vz)$. The forward diffusion process~\cite{dhariwal2021diffusion} gradually perturbs $\vz_0$ into a noised latent $\vz_t$ according to
\begin{equation}
q(\vz_t\!\mid\!\vz_0)=\mathcal{N}\!\big(\vz_t;\sqrt{1-\bar{\alpha}_t}\,\vz_0,\ \bar{\alpha}_t\mathbf I\big),
\end{equation}
where $\bar{\alpha}_t$ is the noise schedule coefficient at timestep $t$.

In text-to-video generation, a prompt $p$ is encoded into a condition $\vc=\mathcal{E}_{\text{text}}(p)$, which guides the denoising process. The training objective minimises the mean squared error between the true noise $\boldsymbol{\epsilon}\sim\mathcal{N}(\mathbf{0}, \mathbf{I})$ and the predicted noise: 
\begin{equation}
\mathbb{E}_{\vz_0,\boldsymbol{\epsilon},t,c}\Big[\;\|\mathbf{\boldsymbol{\epsilon}}-\mathbf{\boldsymbol{\epsilon}}_\theta(\vz,t,\vc)\|^2\;\Big],
\end{equation}
where $\boldsymbol{\epsilon}_\theta(z_t,t,\vc)$ is the noise predictor parameterized by $\theta$.

After training, we generate videos by solving the reverse diffusion ODE~\cite{songscore} using the UniPC sampler~\cite{zhao2023unipc}, a second-order predictor--corrector framework designed for fast and accurate sampling. The ODE is expressed as 
\begin{equation}
\frac{d\vz_t}{dt}=f_\theta(\vz_t,t,\vc),
\end{equation}
where $f_\theta$ is derived from $\boldsymbol{\epsilon}_\theta$ under the probability-flow formulation. To strengthen text guidance, we apply classifier-free guidance (CFG)~\cite{ho2022classifier}:
\begin{equation}
\boldsymbol{\epsilon}_\theta^{\,w}(\vz_t,t,\vc)=\boldsymbol{\epsilon}_\theta(\vz_t,t,\varnothing) + w\big[\boldsymbol{\epsilon}_\theta(\vz_t,t,\vc)-\boldsymbol{\epsilon}_\theta(\vz_t,t,\varnothing)\big],
\end{equation}
where $w \in \mathbb{R}_{\ge 0}$ is the guidance scale and $\varnothing$ denotes the null text prompt.

At each denoising step $t_s \!\to\! t_{s-1}$ with step size $h_s$, UniPC applies a second-order update of the form
\begin{equation}
\label{eq:unipc-solver}
\vz_{t_{s-1}} \;=\; \vz_{t_s} + \tfrac{h_s}{2}\Big(f_\theta(\vz_{t_s}, t_s, \vc) + f_\theta(\tilde{\vz}_{t_{s-1}}, t_{s-1}, \vc)\Big),
\end{equation}
where $\tilde{\vz}_{t_{s-1}} = \vz_{t_s} + h_s f_\theta(\vz_{t_s}, t_s, \vc)$ serves as a predictor. Finally, the terminal latent $\vz_{0}$ is decoded by $\mathcal{D}$ into an $N$-frame RGB video,
$\{\vx^{(i)}\}_{i=1}^N=\mathcal{D}(\vz_{0})$.

\vspace{-1.0em}
\subsection{Latent Reward Model}
\vspace{-0.5em}
To enable intermediate evaluations during video diffusion, we design a latent reward model that can assign quality scores to partially denoised latents without decoding to the video space. We first describe the construction of the training dataset, then detail the architecture of the reward model, and finally present the objective function used to optimise it.

\noindent \textbf{Latent Reward Data Construction.}~Most reward assessors operate on rendered videos and return video-level scores, making it nontrivial to supervise rewards directly on latent representations. Concretely, given a prompt $p$ and the final denoised (``clear'') latent $\vz_0$, the video-level reward vector is obtained on the decoded video 
\begin{equation}
\label{eq:video-reward}
\vr \;=\; \big(r^{\mathrm{VQ}},\, r^{\mathrm{MQ}},\, r^{\mathrm{TA}}\big)^\top \;=\; \mathcal{R}\,\!\big(\mathcal{D}(\vz_0),\, p\big),
\end{equation}
where $\mathcal{D}$ is the decoder, $p$ is the prompt, and $\mathcal{R}$ denotes external verifiers or human annotations for visual quality (VQ), motion quality (MQ), and text alignment (TA)~\cite{liu2025improving}. However, our reward model must evaluate intermediate latents $\vz_t$ extracted along the denoising trajectory. Since direct latent-level ground truth is unavailable, we propose a similarity-grounded credit assignment. Specifically, we ground video-level rewards to intermediate latents by measuring how much an intermediate latent $\vz_t$ ``contributes'' to the final clear latent $\vz_0$ via its similarity to the clear latent. We define a cosine-based similarity, rescaled to $[0,1]$:
\begin{equation}
\label{eq:sim} 
s_t \;=\; \frac{1}{2}\!\left(1 + 
\frac{\langle \vz_t,\, \vz_0\rangle}{\|\vz_t\|_2\,\|\vz_0\|_2}\right)
\;\in\;[0,1].
\end{equation}
The similarity $s_t$ quantifies how close $\vz_t$ is to $\vz_0$ in a task-relevant representation. Finally, we assign latent-level targets by crediting each dimension of the video-level reward proportionally to $s_t$:
\begin{equation}
\label{eq:latent-target}
\tilde{\vr}_t \;=\; s_t \cdot \vr \;=\; \big(s_t\, r^{\mathrm{VQ}},\; s_t\, r^{\mathrm{MQ}},\; s_t\, r^{\mathrm{TA}}\big)^\top.
\end{equation}
This yields the latent reward dataset
\begin{equation}
\label{eq:latent-dataset}
\mathcal{D}_{\mathrm{latent}} \;=\; \big\{\,(\vz_t,\, p,\, \tilde{\vr}_t,\, t)\,:\, t\in\mathcal{T}\,\big\},
\end{equation}
where $\mathcal{T}$ is the set of sampled timesteps.

\noindent \textbf{Model Architecture.}~The reward model takes three inputs: a latent representation $\vz_t\in \mathbb{R}^{F/4\times {C}^\prime \times H/8 \times W/8}$ (where $F$, ${C}^\prime$, $H$, and $W$ denotes video frames number, latent channel, video height, and video width), the denoising step $t$, and the text prompt $p$. These inputs are unified into a token sequence and processed by a transformer-based backbone. \textit{Latent tokens:}~The intermediate latent tensor $\vz_t$ is patchified by a lightweight 3D convolutional encoder, which partitions the spatiotemporal volume into non-overlapping blocks and projects them into embedding vectors. This yields a sequence of video tokens that capture both spatial appearance and temporal motion. \textit{Step embedding:}~The denoising step $t$ is mapped to a learnable embedding vector $\mathbf{e}_t$ through an embedding layer. This embedding is concatenated with the token sequence to provide the model with explicit temporal information about the diffusion process. \textit{Prompt tokens:} The text prompt $p$ is formatted using an instruction-style template designed for reward modeling~\cite{liu2025improving}, and tokenized into instruction tokens. Special query tokens [VQ], [MQ], and [TA] are appended to the sequence to request predictions for visual quality, motion quality, and text–video alignment, respectively. The overall structure is illustrated on the left of Figure~\ref{fig2}.

The unified token sequence is then processed by a transformer-based backbone, which outputs hidden states for the query tokens. Finally, the hidden states corresponding to the three reward query tokens are extracted and projected by a linear layer to obtain scalar scores:
\begin{equation}
\label{eq:reward-head}
\hat{\vr} =\mathrm{Linear}\Big(h^{[\mathrm{VQ}]},\, h^{[\mathrm{MQ}]},\, h^{[\mathrm{TA}]} \Big),
\end{equation}
where $h^{[\cdot]}$ denotes the hidden state of each query token. This yields the predicted rewards $(\hat{r}^{\mathrm{VQ}},\,\hat{r}^{\mathrm{MQ}},\,\hat{r}^{\mathrm{TA}})$.

\noindent \textbf{Training Objective.}~A latent reward model $R_\psi$ is optimised with a combination of a regression loss and a preference loss, showing in \textit{Appendix Algorithm~1}. 

Given the latent tensor $\vz_t$, prompt $p$, and denoising step $t$, the model predicts reward scores $\hat{\vr}=R_\psi(\vz_t,p,t)$ across three dimensions. Since the latent reward dataset $\mathcal{D}_{\mathrm{latent}}=\{(\vz_t,p,\tilde{\vr}_t,t)\}$ already incorporates the similarity weighting, the regression target for each dimension is $\tilde \vr_t^d$, where $d \in\{\text{VQ},\text{MQ},\text{TA}$\}. The regression loss is therefore
\begin{equation}
\label{eq:reg-loss}
\mathcal{L}^d_{\mathrm{reg}} = \big\| \hat{\vr}^d - \tilde \vr_t^d \big\|_2^2.
\end{equation}

While regression provides absolute supervision, it does not enforce relative ordering among candidates. Inspired by reinforcement learning from human feedback (RLHF)~\cite{ouyang2022training}, we introduce a preference loss that encourages the model to predict higher scores for better samples. For each pair $(i,j)$ in a minibatch, we define
\begin{equation}
\label{eq:pref-def}
\Delta \hat{\vr}_{ij}^d = \hat{\vr}_i^d - \hat{\vr}_j^d, \qquad \vy_{ij}^d = \mathbb{I}[\vr_i^d > \vr_j^d],
\end{equation}
where $\vy_{ij}^d$ denotes the ground-truth preference label for pair $(i,j)$ in dimension $d$. The preference loss is then formulated as:
\begin{equation}
\label{eq:pref-loss}
\mathcal{L}^d_{\mathrm{pref}} = \frac{1}{|\mathcal{P}_d|}   \sum_{(i,j)\in\mathcal{P}_d} 
  \log\!\left(1+\exp\!\big(- (2\vy_{ij}^d-1)\,\Delta \hat{\vr}_{ij}^d\big)\right),
\end{equation}
where $\mathcal{P}_d$ is the set of preference pairs. This is equivalent to applying binary cross-entropy on pairwise score differences.

The final loss is a weighted sum over dimensions $d$ and both objectives:
\begin{equation}
\label{eq:total-loss}
\mathcal{L} = \sum_{\mathclap{d\in\{\text{VQ},\text{MQ},\text{TA}\}}} 
\big(\lambda^d_{\mathrm{reg}}\,\mathcal{L}^d_{\mathrm{reg}} 
    + \lambda^d_{\mathrm{pref}}\,\mathcal{L}^d_{\mathrm{pref}}\big).
\end{equation}
We optimise the reward model parameters $\psi$ using stochastic gradient descent. Regression loss anchors the predictions to absolute reward magnitudes, while preference loss shapes the reward landscape to preserve relative orderings, analogous to the role of preference modelling in RLHF.

\vspace{-1.0em}
\subsection{Latent Search with Reward-Guided Resampling and Pruning}
\vspace{-0.5em}
We consider inference-time search as an importance-sampling-inspired procedure on latent trajectories. Standard samplers such as DDIM~\cite{song2020denoising} or UniPC~\cite{zhao2023unipc} follow a single trajectory, which may fall into suboptimal modes. To improve robustness, we maintain a set of $N$ candidate trajectories, inspired by sequential Monte Carlo (SMC) methods~\cite{doucet2001introduction}, and iteratively refine this set during denoising. The algorithm of the proposed latent search method is in \textit{Appendix Algorithm~2.}

\noindent \textbf{Candidate generation.}~At the initial timestep $T$, we generate candidates by perturbing a base Gaussian noise $\vz_T^{(0)}$ with isotropic perturbations $\boldsymbol{\epsilon}_i \sim \mathcal{N}(\mathbf{0}, \mathbf{I})$:
\begin{equation}
\label{eq:candidate-init}
\vz_T^{(i)} \;=\; \sqrt{1-\eta^2}\,\vz_T^{(0)} + \eta\,\boldsymbol{\epsilon}_i, \qquad i=1,\dots,N,
\end{equation}
where $\eta$ controls the candidate diversity. Each candidate is evolved independently with a diffusion sampler.

\noindent \textbf{Reward-guided resampling with uniqueness.}~Let $\mathcal{Z}_{t}=\{(\vz_{t}^{(i)},\sigma_i)\}_{i=1}^{N}$ be the candidate set at step $t$, where $\sigma_i$ is the seed identity of candidate $i$ (determined by its initial noise). At scoring steps $t\in\mathcal{S}$, where $\mathcal{S}$ is the search schedule, we obtain rewards $r_i^{(t)}=R_\psi(\vz_{t}^{(i)},p,t)$ and convert them to normalized weights
\begin{equation}
\label{eq:softmax-weight}
\pi_i^{(t)} \;=\; \frac{\exp(\tau\, r_i^{(t)})}{\sum_{k=1}^{N}\exp(\tau\, r_k^{(t)})}, 
\qquad \sum_i \pi_i^{(t)}=1,
\end{equation}
where $\tau>0$ is a temperature.

We then draw $\vn^{(t)} \sim \mathrm{Multinomial}\big(N;\, \pi_1^{(t)},\dots,\pi_N^{(t)}\big)$ with replacement, and keep the unique seeds only:
\begin{equation}
\label{eq:unique}
\mathcal{I}^{(t)} \;=\; \mathrm{supp}\big(\vn^{(t)}\big) \;=\; \{\, i \mid n_i^{(t)} > 0 \,\},
\qquad
\mathcal{Z}_{t}^{+} \;=\; \big\{\,(\vz_{t}^{(i)},\sigma_i)\,:\, i\in\mathcal{I}^{(t)}\big\}.
\end{equation}
The uniqueness operator, $\mathrm{supp}(\cdot)$, removes multiplicities (duplicates of the same seed) and thus avoids wasting compute on identical trajectories, in which the survival probability of candidate $i$ after uniqueness is $1-\big(1-\pi_i^{(t)}\big)^{N}$. This increases monotonically with its weight $\pi_i^{(t)}$.

Between two scoring steps, there are many intermediate denoising updates, during which candidates evolve independently via the diffusion sampler.

\noindent \textbf{Cumulative weighting and final pruning.}~We accumulate evidence across scoring steps via an additive criterion $c_i^{(t)} \;=\; c_i^{(t-1)} + \pi_i^{(t)}$, in which $c_i^{(0)}=0$. At the final scheduled step $t^{\prime}=\max(\mathcal{S})$, if multiple candidates remain, we prune by selecting the seed with the highest cumulative weight:
\begin{equation}
\label{eq:final-prune}
i^\star \;=\; \arg\max_{\,i\in\mathcal{I}^{(t^{\prime})}} \; c_i^{(t^{\prime})}, \qquad \vz_0 \;=\; \vz_0^{(i^\star)}.
\end{equation}
The surviving latent $\vz_0$ is then decoded into the final video.

Overall, LatSearch follows the spirit of importance sampling: multiple candidate trajectories are proposed in latent space, weighted according to reward scores that act as surrogate importance factors, and resampled to balance exploration and exploitation. The final pruning step selects the most consistently high-reward trajectory, yielding the decoded video.

\vspace{-1em}
\section{Experiments}
\vspace{-0.5em}
In this section, we evaluate the efficacy of our LatSearch through extensive experiments on a large-scale text-to-video generation task. We will first detail the implementations and then compare our method to other state-of-the-art inference-time scaling methods for video generation. We finally present the ablation studies. Additional qualitative comparisons and detailed experimental analyses are provided in the \textit{Appendix}.

\vspace{-1.5em}
\subsection{Experimental Settings}
\vspace{-0.5em}
\textbf{Implementations.}
We adopt Qwen2-VL-3B~\cite{wang2024qwen2} as the backbone for our reward model, owing to its strong performance on multimodal understanding tasks and its suitability for video–language alignment. To construct the latent reward pairs dataset, we first sample 1,000 text prompts that are non-overlapping with VBench-2.0. Using these prompts, we generate 5,000 videos with different random seeds, while also storing the corresponding latents at selected timesteps and their similarity to the final denoised latent. The dataset is partitioned into 80\% for training and 20\% for testing. For latent reward model training, we initialise the learning rate at 1e-4 and reduce it to 1e-5 at the 10th epoch. The model is trained with a batch size of 4 and stopped at the 15th epoch. Both the regression loss and preference loss are weighted equally with a coefficient of 1.0. We adopt Wan2.1-1.3B~\cite{wan2025wan} as the baseline video generative model, where inference steps are set to 50 and the CFG scale to 5.0 as suggested in~\cite{wan2025wan}. Following~\cite{he2025scaling}, each generated video consists of 33 frames at a resolution of $832\times480$. For LatSearch, the scoring schedule is applied at timesteps 10, 15, and 20. All experiments are conducted on NVIDIA A100 GPUs.

\noindent \textbf{Evaluation.}~To evaluate the performance of each method, we use VBench-2.0~\cite{zheng2025vbench}, a comprehensive benchmark that automatically evaluates video generative models for their intrinsic faithfulness. Specifically, VBench-2.0 assesses video performance across five emerging dimensions beyond superficial faithfulness: Human Fidelity, Controllability, Creativity, Physics, and Commonsense. The scores range from 0 to 100, with a higher score indicating better performance in the corresponding aspects. For each noise prior, we generate 3,860 videos for VBench-2.0 evaluation. Detailed implementation settings and evaluation protocols are provided in \textit{Appendix Sec. B}.

\vspace{-1.5em}
\subsection{Comparisons to State of the Art}
\vspace{-0.5em}
To validate the effectiveness of our method for video inference-time scaling, we first compare it against existing inference-time scaling approaches on VBench-2.0. Specifically, we evaluate against methods that optimise the initial noise, including FreeInit~\cite{wu2024freeinit} and FreqPrior~\cite{yuan2025freqprior}, as well as search-based approaches, including VideoReward~\cite{liu2025improving} and EvoSearch~\cite{he2025scaling}. We report results across the five evaluation dimensions of VBench-2.0, along with their averaged scores, and also provide the inference time of each method. Full per-dimension evaluation results are reported in \textit{Appendix Sec. C.4}.

\begin{table}[!t]
\begin{center}
\caption{Comparison of inference-time scaling methods for video generation on VBench-2.0. The table includes both optimisation-based and search-based approaches. $^\dagger$ indicates the use of DPM-Solver++~\cite{lu2025dpm}. \textbf{Bold} numbers indicate the best results, while \underline{underlined} numbers indicate the second-best. $\uparrow$ denotes performance degradation and $\downarrow$ denotes performance increase.}
\vspace{-1em}
\label{tab1}
\setlength{\tabcolsep}{3pt}
\resizebox{\columnwidth}{!}{%
\begin{tabular}{@{}c|ccccc|>{\columncolor{gray!15}}c@{}|>{\columncolor{gray!15}}c}
\toprule
Methods                            & Creativity & Commonsense & Controllability & Human Fidelity & Physics & Average       & Inference Time (s)   \\ 
\midrule
Baseline                           & 53.81      & 55.63       & 21.99           & 82.11          & 45.98   & $51.90$         & $77.21\pm0.26$          \\ 
\midrule
FreeInit {[}ECCV'24{]}             & 46.80      & \underline{59.41} & 22.03 & \underline{85.22} & 35.65     & $49.82_{(-2.08)}\downarrow$ & $308.87\pm0.22_{(\times4.00)}$ \\
FreqPrior {[}ICLR'25{]}            & 53.70      & 55.61       & 20.35           & 83.61        & 38.60      & $50.37_{(-1.53)}\downarrow$ & $142.46\pm0.81_{(\times\textbf{1.85})}$ \\
VideoReward {[}NeurIPS'25{]}       & 55.07      & 57.91       & 23.15           & 82.21        & 45.67      & $52.80_{(+0.90)}\uparrow$   & $283.63\pm2.42_{(\times3.67)}$ \\
EvoSearch$^\dagger$ {[}arXiv'25{]} & \textbf{59.25} & \textbf{61.09} & \textbf{26.18} & \textbf{86.73} & 41.80 & $\underline{55.01}_{(+3.11)}\uparrow$ & $783.76\pm3.15_{(\times10.15)}$ \\ 
\midrule
\textbf{LatSearch (Ours)}           & \underline{58.12} & 59.37  & 22.69 & 82.59  & \underline{46.44} & $53.84_{(+1.94)}\uparrow$ & $182.43\pm6.53_{(\times\textbf{2.36})}$ \\
\textbf{LatSearch$^\dagger$ (Ours)} & 57.53 & 57.36  & \underline{23.96}  & 84.01 & \textbf{53.41} & $\textbf{55.25}_{(\textbf{+3.35})}\uparrow$ & $164.41\pm4.79_{(\times\textbf{2.13})}$ \\ 
\bottomrule
\end{tabular}
}
\vspace{-1em}
\end{center}
\end{table}
\begin{table}[!t]
\vspace{-1em}
\begin{center}
\caption{Comparison of varied search budgets $N$ on VBench-2.0.}
\vspace{-2em}
\label{tab2}
\setlength{\tabcolsep}{3pt}
\resizebox{\columnwidth}{!}{%
\begin{tabular}{@{}c|ccccc|>{\columncolor{gray!15}}c@{}|>{\columncolor{gray!15}}c}
\toprule
Search Budget $N$ & Creativity & Commonsense & Controllability & Human Fidelity & Physics & Average       & Inference Time (s)   \\ 
\midrule
Baseline                         & 53.81      & 55.63       & 21.99           & 82.11          & 45.98   & $51.90$         & $77.21\pm0.26$          \\ 
\midrule
4  & 57.70 & 54.70  & 22.00 & 83.63  & 46.03 & $52.81_{(+0.91)}$& $132.71 \pm 2.55_{(\times 1.72)}$\\
6  & 58.12& \textbf{59.37}  & \textbf{22.69} & 82.59  & 46.44& $53.84_{(+1.94)}$& $182.43\pm6.53_{(\times2.36)}$\\
8 & \textbf{58.47}& 58.87& 22.26& \textbf{84.00}& \textbf{47.05}& $\textbf{54.13}_{(+2.23)}$& $225.56\pm15.57
_{(\times2.92)}$\\ 
\bottomrule
\end{tabular}
}
\vspace{-3em}
\end{center}
\end{table}

The main comparison results are reported in Table~\ref{tab1}. For methods that optimise the initial noise, although they do not significantly increase computational cost, the absence of an effective verifier limits their effectiveness: simply extracting temporal features from the latent space and feeding them back into the initial noise fails to improve video generation quality. In contrast, search-based approaches rely on output reward optimisation, either via greedy or evolutionary algorithms. While these methods improve quality, they require several times more search time than the baseline, undermining efficiency. Our approach differs in that it evaluates the latent states directly at intermediate steps and employs a probabilistic sampling strategy to retain promising candidate seeds adaptively. As shown in Table~\ref{tab1}, our best configuration improves video quality by 3.35\% over the baseline, while requiring only 2.13$\times$ more computation. Compared to FreqPrior, under the same computational budget, our method achieves a 2.44\% higher quality score. When compared with EvoSearch, although our method achieves only a modest 0.24\% gain in quality, it is 4.77$\times$ faster. Additional qualitative comparisons are provided in \textit{Appendix Sec. D}.

\vspace{-1.5em}
\subsection{Ablation Analysis}
\vspace{-0.5em}
We conduct ablation studies to better understand the key design choices of LatSearch. Additional analyses on credit assignment strategies, preference loss, temperature sensitivity, and scoring schedules are provided in \textit{Appendix Sec. C.1 and C.2}, together with a human preference study in \textit{Appendix Sec. C.3}.

\noindent \textbf{Effect of Search Budget $\textbf{\textit{N}}$.}~We evaluate how the number of candidate trajectories $N$ influences the performance of LatSearch. As reported in Table~\ref{tab2}, increasing the search budget leads to consistent improvements across all VBench-2.0 dimensions. Moving from $N=4$ to $N=6$ yields noticeable gains, particularly in Creativity, Commonsense, and Human Fidelity, demonstrating that maintaining a larger pool of latent trajectories enables the search mechanism to explore richer solution paths. Further increasing the budget to $N=8$ continues to improve performance, indicating that the proposed latent-level scoring and resampling procedure can effectively utilise additional candidates when available. However, the benefits gradually saturate with growing $N$. Although $N=8$ produces the strongest results, its marginal improvement over $N=6$ is smaller compared to the earlier jump from $N=4$. Importantly, the increase in computational cost remains moderate because evaluations are performed directly in latent space, avoiding repeated decoding into video space. This property allows LatSearch to scale gracefully with search budget while remaining computationally practical.

\noindent \textbf{Generalisation Across Backbones.}~To assess the generality of our approach, we evaluate LatSearch across different video diffusion backbones. Since LatSearch operates in latent space and only requires reward evaluation of intermediate denoising states, it does not depend on architecture-specific components. We first apply LatSearch to the larger Wan2.1-14B backbone. As shown in Table~\ref{tab3}, LatSearch improves the average VBench-2.0 score from 52.58 to 53.61, with gains across all five evaluation dimensions. This indicates that latent reward guidance remains effective when scaling to a stronger generation backbone.

To further evaluate whether LatSearch generalises beyond the Wan model family, we apply it to two CogVideoX backbones, CogVideoX-2B and CogVideoX-5B. As shown in Table~\ref{tab3}, LatSearch improves the average score from 45.13 to 48.14 on CogVideoX-2B, yielding a gain of +3.01, and from 48.13 to 50.21 on CogVideoX-5B, yielding a gain of +2.08. Although the improvements vary across individual dimensions, the consistent average gains across both CogVideoX models demonstrate that LatSearch is not tied to a specific video diffusion architecture and can provide effective scaling across different model families.

\begin{figure}[!t]
    \centering 
    \includegraphics[width=\textwidth]{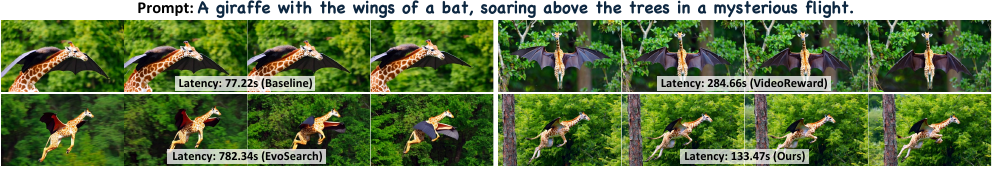}
    \vspace{-2em}
    \caption{Qualitative comparison with search-based video generation methods. VideoReward achieves strong semantic alignment but suffers from poor temporal dynamics. EvoSearch improves both semantics and dynamics, yet requires heavy search cost. Our LatSearch reaches comparable quality to EvoSearch while being nearly $5\times$ faster. Results are better viewed with zoom-in.} 
    \label{fig3}
    \vspace{-1em}
\end{figure}
\begin{table}[!t]
\begin{center}
\caption{Video generation results across different backbones.}
\vspace{-1em}
\label{tab3}
\setlength{\tabcolsep}{3pt}
\resizebox{0.9\columnwidth}{!}{%
\begin{tabular}{@{}c|ccccc|>{\columncolor{gray!15}}c}
\toprule
Methods & Creativity & Commonsense & Controllability & Human Fidelity & Physics & Average \\ 
\midrule
Wan2.1-14B            & 55.21      & 56.18       & 21.79        & 89.74          & 39.96   & 52.58 \\ 
+ LatSearch  & \textbf{56.28} & \textbf{57.64} & \textbf{22.52} & \textbf{91.45}  & \textbf{40.17} & $\textbf{53.61}_{(+1.03)}$ \\
\midrule
CogVideoX-2B & 40.97 & 37.48 & \textbf{26.92} & \textbf{81.56} & 38.74 & 45.13 \\
+ LatSearch & \textbf{49.59} & \textbf{45.08} & 25.94 & 77.11 & \textbf{42.94} & $\textbf{48.14}_{(+3.01)}$ \\ 
\midrule
CogVideoX-5B & 42.52 & 45.64 & 26.53 & 82.22 & \textbf{43.74} & 48.13 \\
+ LatSearch & \textbf{50.91} & \textbf{48.03} & \textbf{27.29} & \textbf{83.08} & 41.72 & $\textbf{50.21}_{(+2.08)}$ \\
\bottomrule
\end{tabular}
}
\vspace{-2em}
\end{center}
\end{table}
\begin{table}[!t]
\begin{center}
\caption{Comparison of VBench-2.0 results under different search strategies and latent reward model settings. A beam-search–style strategy is used for the variant without our PGRP. PL denotes a latent reward model trained with preference loss.}
\vspace{-1.7em}
\label{tab4}
\setlength{\tabcolsep}{3pt}
\resizebox{\columnwidth}{!}{%
\begin{tabular}{@{}c|c|c|ccccc|>{\columncolor{gray!15}}c|>{\columncolor{gray!15}}c}
\toprule
Methods     &  PL          &  RGRP        & Creativity    & Commonsense   & Controllability & Human Fidelity & Physics       & Average       & Inference Time (s)   \\ 
\midrule
Baseline    & $\times$     & $\times$     & 53.81         & 55.63         & 21.99           & 82.11          & 45.98         & 51.90         & 77.21 ($\pm 0.26$)   \\  
\midrule
            & $\checkmark$ & $\times$     & 56.63$_{(+2.82)}$ & 57.07$_{(+1.44)}$ & 22.34$_{(+0.35)}$ & 82.54$_{(+0.43)}$ & 43.16$_{(-2.82)}$ & 52.35$_{(+0.45)}$ & $171.23 (\pm 1.42)_{(\times2.22)}$ \\
LatSearch & $\times$     & $\checkmark$  & 56.44$_{(+2.64)}$ & 58.51$_{(+2.89)}$ & 22.28$_{(+0.29)}$   & \textbf{84.33}$_{(+2.22)}$ & 45.54$_{(-0.45)}$ & 53.42$_{(+1.52)}$ & $182.43 (\pm 6.53)_{(\times2.36)}$ \\
            & $\checkmark$ & $\checkmark$ & \textbf{58.12}$_{(+4.31)}$ & \textbf{59.37}$_{(+3.74)}$ & \textbf{22.69}$_{(+0.70)}$ & 82.59$_{(+0.48)}$ & \textbf{46.44}$_{(+0.46)}$ & \textbf{53.84}$_{(+1.94)}$ & $182.43 (\pm 6.53)_{(\times2.36)}$ \\ 
\bottomrule
\end{tabular}
}
\vspace{-3em}
\end{center}
\end{table}

\noindent \textbf{Effectiveness of the RGRP.}~We compare RGRP with a beam search–style approach that relies solely on cumulative reward scores. Concretely, both methods follow the same setting of scoring latent candidates at scheduled denoising timesteps. The difference is that the beam search baseline retains a fixed number of seeds purely based on accumulated rewards, while our RGRP method incorporates probabilistic resampling and pruning guided by reward signals. As shown in Table~\ref{tab4}, RGRP achieves consistently better results across most of the five evaluation dimensions. On average, it improves video quality by 0.42\% compared to the beam search–style baseline, while incurring only a marginal increase in computation. These results highlight the benefit of probabilistic selection in balancing exploration and exploitation, leading to more robust search outcomes than deterministic reward accumulation. This demonstrates that RGRP mitigates overfitting to accumulated reward signals and promotes more diverse yet high-quality candidate selection.

\noindent \textbf{Different Credit Assignment Strategies.}~To investigate how reward signals should be propagated across denoising steps, we evaluate several credit assignment strategies, including Uniform, Exponential, L2 Error, and Cosine Similarity weighting. As shown in Table ~\ref{tab5}, Cosine Similarity achieves the best overall performance, improving the VBench-2.0 average score from 51.90 to 53.84. This strategy notably enhances Creativity and Commonsense while maintaining stable performance across Controllability and Physical plausibility. In contrast, Uniform and Exponential weighting provide only marginal gains, suggesting that naïvely distributing rewards across timesteps is insufficient. We hypothesise that Cosine Similarity better aligns latent updates with semantic consistency, enabling more effective guidance during latent reward model training. Additional empirical approximation-error analysis is provided in \textit{Appendix Sec.~C.1}.

\begin{table}[!t]
\begin{center}
\caption{VBench-2.0 evaluation results under different credit assignment strategies.}
\vspace{-1em}
\label{tab5}
\setlength{\tabcolsep}{3pt}
\resizebox{0.97\columnwidth}{!}{%
\begin{tabular}{@{}c|ccccc|>{\columncolor{gray!15}}c}
\toprule
Methods    & Creativity & Commonsense & Controllability & Human Fidelity & Physics & Average \\ 
\midrule
Baseline           & 53.81 & 55.63 & 21.99 & 82.11 & 45.98 & 51.90 \\ 
\midrule
Uniform            & 54.98 & 52.40 & 21.44 & 84.98 & 47.75 & $ 52.31_{(+0.41)}$ \\ 
Exponential        & 54.50 & 52.09 & 21.63 & 83.03 & 49.30 & $ 52.11_{(+0.21)}$ \\ 
L2 Error           & 55.78 & 56.64 & 22.20 & 82.12 & 46.95 & $ 52.74_{(+0.84)}$ \\ 
Cosine Similarity  & 58.12 & 59.37 & 22.69 & 82.59 & 46.44 & $ \textbf{53.84}_{(+1.94)}$ \\ 
\bottomrule
\end{tabular}
}
\vspace{-2em}
\end{center}
\end{table}
\begin{table}[!t]
\begin{center}
\caption{Results of LatSearch across different video lengths and resolutions.}
\vspace{-2em}
\label{tab_length_resolution}
\setlength{\tabcolsep}{2pt}
\resizebox{\columnwidth}{!}{%
\begin{tabular}{c|c|ccccc|>{\columncolor{gray!15}}c}
\toprule
Setting & Methods & Creativity & Commonsense & Controllability & Human Fidelity & Physics & Average \\ 
\midrule
\midrule
\multirow{2}{*}{\begin{tabular}[c]{@{}c@{}}2s\\(32-frame)\end{tabular}}
& Wan-1.3B & 53.81 & 55.63 & 21.99 & 82.11 & 45.98 & 51.90 \\
& + LatSearch & \textbf{58.12} & \textbf{59.37} & \textbf{22.69} & \textbf{82.59} & \textbf{46.44} & $\textbf{53.84}_{(+1.94)}$ \\ 
\midrule
\multirow{2}{*}{\begin{tabular}[c]{@{}c@{}}3s\\(48-frame)\end{tabular}}
& Wan-1.3B & 54.41 & \textbf{54.10} & 21.11 & 81.72 & 41.97 & 50.66 \\
& + LatSearch & \textbf{55.00} & 53.23 & \textbf{21.70} & \textbf{82.18} & \textbf{47.23} & $\textbf{51.87}_{(+1.21)}$ \\ 
\midrule
\multirow{2}{*}{\begin{tabular}[c]{@{}c@{}}4s\\(64-frame)\end{tabular}}
& Wan-1.3B & \textbf{53.35} & 50.96 & 22.47 & 79.03 & \textbf{45.95} & 50.35 \\
& + LatSearch & 52.87 & \textbf{56.16} & \textbf{23.18} & \textbf{80.23} & 43.66 & $\textbf{51.22}_{(+0.87)}$ \\ 
\midrule
\multirow{2}{*}{\begin{tabular}[c]{@{}c@{}}5s\\(80-frame)\end{tabular}}
& Wan-1.3B & \textbf{54.88} & 53.25 & \textbf{23.60} & 77.38 & 39.88 & 49.80 \\
& + LatSearch & 54.01 & \textbf{55.87} & 22.59 & \textbf{80.74} & \textbf{42.65} & $\textbf{51.17}_{(+1.37)}$ \\ 
\midrule
\midrule
\multirow{2}{*}{\begin{tabular}[c]{@{}c@{}}480p\\(832$\times$480)\end{tabular}}
& Wan-1.3B & 53.81 & 55.63 & 21.99 & 82.11 & 45.98 & 51.90 \\
& + LatSearch & \textbf{58.12} & \textbf{59.37} & \textbf{22.69} & \textbf{82.59} & \textbf{46.44} & \textbf{53.84 (+1.94)} \\ 
\midrule
\multirow{2}{*}{\begin{tabular}[c]{@{}c@{}}720p\\(1280$\times$720)\end{tabular}}
& Wan-1.3B & 41.52 & \textbf{59.07} & \textbf{22.29} & \textbf{81.44} & 37.67 & 48.40 \\
& + LatSearch & \textbf{42.12} & 57.63 & 22.09 & 81.12 & \textbf{43.03} & \textbf{49.20 (+0.80)} \\ 
\bottomrule
\end{tabular}
}
\vspace{-3em}
\end{center}
\end{table}
\noindent \textbf{Scalability to Video Lengths and Resolutions.}~We also examine LatSearch under different video lengths and resolutions. Table~\ref{tab_length_resolution} reports results for 2s/32-frame, 3s/48-frame, 4s/64-frame, and 5s/80-frame generation, as well as 480p and 720p resolutions. LatSearch improves the average VBench-2.0 score by +1.94, +1.21, +0.87, and +1.37 for 2s, 3s, 4s, and 5s videos, respectively. It also improves performance at both 480p and 720p, with gains of +1.94 and +0.80. These results suggest that LatSearch remains effective beyond the default evaluation setting, although the gains become smaller for more challenging longer-duration and higher-resolution generation.

\begin{wrapfigure}{}{0.48\textwidth} 
    \vspace{-2.3em}
    \centering
    \includegraphics[width=0.45\textwidth]{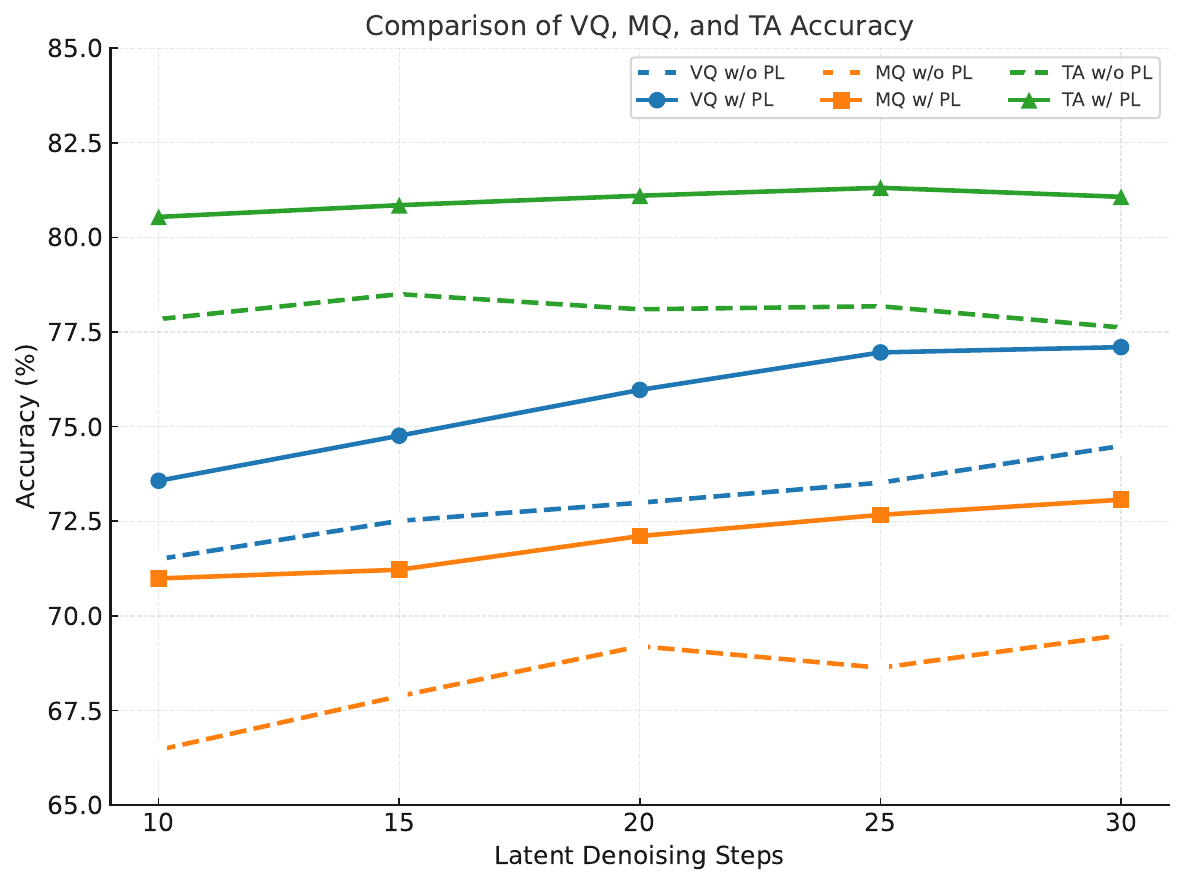}
    \vspace{-0.7em}
    \caption{Comparison of VQ, MQ, and TA accuracy across different loss function settings and denoising steps.}
    \vspace{-2.2em}
    \label{fig3}
\end{wrapfigure}

\noindent \textbf{Effectiveness of the Preference Loss.} To validate the effectiveness of incorporating preference loss into the latent reward model, we conduct experiments from two perspectives: (1) the consistency between the latent reward and the video-level verifier, and (2) the improvement in video generation quality under our proposed LatSearch framework. For the first perspective, we construct approximately 446K latent pairs in the test set, with preferences computed from the corresponding video scores and similarity to the final clean latent. As shown in Figure~\ref{fig3}, compared with regression loss, preference loss improves the alignment accuracy by 3.1\%, 2.65\%, 2.96\%, 3.54\%, and 3.21\% at denoising steps of 10, 15, 20, 25, and 30, respectively. This demonstrates the effectiveness of preference learning. Furthermore, we directly validate the preference loss within our latent search method without RGRP. As shown in Table~\ref{tab4}, the use of preference loss leads to a 1.07\% improvement in video quality. This further highlights the effectiveness of our search algorithm: as the latent reward model becomes stronger, our method consistently achieves better performance. These results confirm that preference-based supervision is not only beneficial at the model level but also translates into measurable improvements in downstream video generation.

\noindent \textbf{Inference Time Comparison.}~Table~\ref{tab6} presents a detailed runtime breakdown across different methods. VideoReward and EvoSearch incur substantial computational overhead due to repeated full video decoding and the maintenance of multiple candidate trajectories, resulting in significantly increased DiT and decoder costs. For example, EvoSearch requires $756.66$ seconds of DiT computation and $31.99$ seconds of decoding on average, leading to a total runtime of $790.57$ seconds.

In contrast, LatSearch operates directly in the latent space, keeping decoder usage nearly identical to the baseline ($1.88$ vs.\ $1.85$ seconds) and introducing only a lightweight latent reward evaluation cost with $1.03$ seconds. This efficiency stems from the low overhead of latent-space operations: each DiT forward step takes only $1.35$ seconds, while latent reward evaluation requires $0.84$ seconds per latent, substantially cheaper than full video decoding. As a result, LatSearch achieves the best VBench-2.0 performance at 55.25\% while requiring only $159.02$ seconds in total runtime, which is nearly 5$\times$ faster than EvoSearch and substantially more efficient than VideoReward, requiring $277.39$ seconds.

\begin{table}[!t]
\centering
\caption{Inference time comparison across different methods.}
\vspace{-1em}
\label{tab6}
\resizebox{\columnwidth}{!}{
\begin{tabular}{c|c|c|c|c|c}
\toprule
Methods & DiT Time ($\downarrow$) & Decoder Time ($\downarrow$) & Reward Time ($\downarrow$) & Total Time ($\downarrow$) & VBench-2.0 Results ($\uparrow$) \\
\midrule
Baseline      & $67.46 \pm 0.71$   & $1.85 \pm 0.26$  & $0$             & $69.31 \pm 0.76$   & 51.90 \\
VideoReward   & $266.71 \pm 0.40$  & $10.12 \pm 0.54$ & $0.56 \pm 0.33$ & $277.39 \pm 0.75$  & 52.80 \\
EvoSearch     & $756.66 \pm 1.27$  & $31.99 \pm 0.96$ & $1.92 \pm 1.04$ & $790.57 \pm 1.90$  & 55.01 \\
LatSearch     & $156.11 \pm 5.11$  & $1.88 \pm 0.24$  & $1.03 \pm 0.06$ & $159.02 \pm 5.12$  & \textbf{55.25} \\
\bottomrule
\end{tabular}
}
\vspace{-2em}
\end{table}

\vspace{-1em}
\section{Conclusion}
\vspace{-0.5em}
We have presented LatSearch, a new inference-time scaling framework for video diffusion that addresses the limitations of existing methods, which either impose priors on initial noise or evaluate only final decoded videos. By introducing a latent reward model for intermediate evaluation and integrating it with Reward-Guided Resampling and Pruning (RGRP), LatSearch achieves both higher video quality and improved efficiency. Specifically, it attains 2.44\% higher fidelity quality under equal compute cost, and is $4.77\times$ faster when achieving the same fidelity quality, compared to state-of-the-art methods. Experiments on VBench-2.0 demonstrate consistent improvements across diverse evaluation dimensions, highlighting latent reward guidance as a promising direction for scalable and efficient video generation.

\vspace{-1em}
\section{Limitations \& Future Work}
\vspace{-0.5em}
\textbf{Limitations.}~While LatSearch demonstrates consistent performance improvements and considerable inference-time savings, several limitations remain. Firstly, our resampling-and-pruning procedure is inspired by Sequential Monte Carlo methods, yet theoretical convergence guarantees are difficult to establish due to the learned and approximate nature of the latent reward model. Consequently, although empirically effective, we do not claim formal optimality of the search procedure. Secondly, we rely on cosine-similarity weighting to transform video-level rewards into latent-level supervision. Although ablations show this strategy performs best among tested alternatives, it remains an approximation of true semantic contribution. More principled or learned temporal credit-assignment functions may further improve latent-reward consistency. 

\noindent \textbf{Future Work.}~Firstly, developing a lightweight temporal similarity estimator, potentially contrastive or self-supervised, could provide a more accurate credit-assignment mechanism, directly addressing the current approximation bottleneck. Secondly, since two dimensions of our reward model (visual quality and motion quality) are modality-agnostic, LatSearch can be extended to audio–video generation, video editing, and instruction-guided transformations, by replacing the text-alignment objective with a task-specific alignment module.

\noindent \textbf{Acknowledgments.}~This research utilised Queen Mary’s Apocrita HPC facility, supported by QMUL Research-IT. Zengqun Zhao was funded by Queen Mary Principal’s PhD
Studentships. This work was supported by Huawei Technologies Germany (Project 19655832: Uncertainty Aware Data Attribution Of Vision-Language Models).

\clearpage
\bibliographystyle{splncs04}
\bibliography{reference}

\clearpage
\appendix
\counterwithin{figure}{section}
\counterwithin{table}{section}

\vspace{-2em}
\section*{Appendix}
\vspace{-0.5em}
This appendix provides supplementary materials that support the main paper. 
We first present the detailed algorithms of the proposed latent reward model training and the latent reward-guided inference-time search procedure. 
We then describe the experimental settings in greater detail, including the calculation of VBench-2.0 metrics and the implementation details of all compared methods. 
In addition, we provide extended experimental results, covering ablation studies of the latent reward model, sensitivity analyses of key search hyperparameters, human preference evaluation, and full per-dimension VBench-2.0 results across different methods and model backbones. 
Finally, we include additional qualitative comparisons to further illustrate the advantages of LatSearch.

\vspace{0.5em}
\noindent\textbf{Contents of this appendix.}
\begin{itemize}
    \item \textbf{Section A: Algorithms.} Detailed pseudocode for latent reward model training and latent reward-guided inference-time search.
    \item \textbf{Section B: Detailed Experimental Settings.} Metric definitions, implementation details, and evaluation protocols.
    \item \textbf{Section C: Additional Experimental Results.} Ablation studies, hyperparameter sensitivity analysis, human preference study, and full VBench-2.0 results.
    \item \textbf{Section D: Qualitative Comparisons.} Additional visual examples comparing the baseline model and LatSearch.
\end{itemize}

\vspace{-2em}
\section{\bfseries Algorithms}
\vspace{-1em}
In this section, we present the detailed algorithms for latent reward model training (Alg.~\ref{alg:train-reward}) and latent reward-guided inference-time search (Alg.~\ref{alg:latsearch}).
\begin{algorithm}[]
\footnotesize
\DontPrintSemicolon
\caption{Training the Latent Reward Model}
\label{alg:train-reward}
\KwIn{
Training set $\{(\vz, p, r^{\text{VQ}}, r^{\text{MQ}}, r^{\text{TA}}, s_t, t)\}$ (latent tensors $\vz$, prompt $p$, reward labels $r^d$, \\
latent similarity $s$, denoising step $t$). Model $R_\psi$; optimizer $\mathcal{O}$; loss weights $\lambda^d_{\text{reg}}, \lambda^d_{\text{pref}}$.
}
\KwOut{Updated reward model parameters $\psi$.}

\For{epoch $=1 \dots E$}{
  \For{batch $\in$ dataloader}{
    $\hat{r} = R_\psi(\vz, p, t)$ \tcp*{\small Predict [$\hat{r}^{\text{VQ}},\hat{r}^{\text{MQ}},\hat{r}^{\text{TA}}$]}
    
    \For{each dimension $d$}{
      $\mathcal{L}^d_{\text{reg}} \gets \| \hat{r}^d - r^d \|^2$ \tcp*{\small Regression loss}
      For all pairs $(i,j)$: \\
      \Indp
      $\Delta \hat{r}_{ij}^d \gets \hat{r}_i^d - \hat{r}_j^d,\quad 
       y_{ij}^d \gets \mathbb{I}[r_i^d > r_j^d]$\;
      $\mathcal{L}^d_{\text{pref}} \gets \text{BCEWithLogits}(\Delta \hat{r}_{ij}^d, y_{ij}^d)$ \tcp*{\small Preference loss}
      \Indm
    }
    $\mathcal{L} \gets \sum_d \big( \lambda^d_{\text{reg}} \mathcal{L}^d_{\text{reg}} + \lambda^d_{\text{pref}} \mathcal{L}^d_{\text{pref}} \big)$ \tcp*{\small Combined loss}
    $\mathcal{O}.zero\_grad()$; Backpropagate $\nabla_\psi \mathcal{L}$; $\mathcal{O}.step()$ \tcp*{\small Optimization}
  }
}
\end{algorithm}
\begin{algorithm}[t]
\DontPrintSemicolon
\caption{Latent Reward-Guided Inference Time Search}
\label{alg:latsearch}
\KwIn{
Prompt $p$; frame number $F$; resolution $(H, W)$; sampling steps $T$; guidance scale $w$; 
search schedule $\mathcal{S}$; number of candidates $N$; noise mixing $\eta$; reward verifier $\hat\gR_{\psi}$.
}
\KwOut{Generated video $\mathbf{x}$.}

\BlankLine
\textbf{Initialization:} \\
Sample base noise $\vz_T^{(0)} \sim \mathcal{N}(0, I)$; sample perturbations $\boldsymbol{\epsilon}_i \sim \mathcal{N}(0,I)$ for $i=1,\dots,N-1$; \\
Construct candidate set $\vz_T^{(i)} \gets \sqrt{1-\eta^2}\,\vz_T^{(0)} + \eta\,\boldsymbol{\epsilon}_i$; \\
Initialize weights $w_i \gets 1/N$ and cumulative weights $c_i \gets 0$; \\
Instantiate $N$ independent diffusion schedulers.

\BlankLine
\For{$j=1$ \KwTo $T$}{
  \For{$i=1$ \KwTo $N$}{
    $\hat{\boldsymbol{\epsilon}} \gets \epsilon_\theta(\vz_t,t,\varnothing) + w\big[\epsilon_\theta(\vz_t,t,\vc)-\epsilon_\theta(\vz_t,t,\varnothing)\big]$ \tcp*{\small CFG}

    $\vz_{t_{j-1}}^{(i)} \gets \text{SamplerStep} (\vz_{t_j}^{(i)}, \hat{\boldsymbol{\epsilon}})$ \tcp*{\small One sampler step}
  }
  \tcp{\small Evaluate and resample candidates}
  \If{$j \in \mathcal{S}$}{
    $r_i \gets \mathcal{R}(\vz_{t_{j-1}}^{(i)}, p, t_j)$ for $i=1,\dots,N$\;
    $w_i \gets \text{Softmax}(r_i)$; \quad $c_i \gets c_i + w_i$\;
    
    Resample indices $\mathcal{I} \sim \text{Multinomial}(w)$\;
    Retain unique $\{\vz^{(i)}, c_i\}_{i \in \mathcal{I}}$ and their schedulers; set $N \gets |\mathcal{I}|$\;
    \tcp{\small Final pruning}
    \If{$j = \max(\mathcal{S})$}{ 
      $i^\star \gets \arg\max_i c_i$; retain only $\vz^{(i^\star)}$ and scheduler; set $N \gets 1$ 
    } 
  }
}

\BlankLine
\textbf{Decode:} $\mathbf{x} \gets \text{VAE.decode}(\vz_0)$\;

\Return $\mathbf{x}$\;
\end{algorithm}

\vspace{-1.5em}
\section{Detailed Experimental Settings}
\vspace{-0.5em}
\textbf{Calculation of VBench-2.0 Matrices.}~We report evaluation results based on the VBench-2.0 metrics. Specifically, each high-level score is computed as the mean of several fine-grained dimensions:
\begin{itemize}
    \item \textbf{Creativity} Score = average of Diversity and Composition.
    \item \textbf{Commonsense} Score = average of Motion Rationality and Instance Preservation.
    \item \textbf{Controllability} Score = average of Dynamic Spatial Relationship, Dynamic Attribute, Motion Order Understanding, Human Interaction, Complex Landscape, Complex Plot, and Camera Motion.
    \item \textbf{Human Fidelity} Score = average of Human Anatomy, Human Identity, and Human Clothes.
    \item \textbf{Physics} Score = average of Mechanics, Thermotics, Material, and Multi-View Consistency.
\end{itemize}
Finally, the Total Score is obtained by averaging the above five high-level scores: Creativity, Commonsense, Controllability, Human Fidelity, and Physics.

\noindent \textbf{Implementations of Compared Methods.}~We compare our approach with two types of baselines: (i) noise-optimisation methods, including FreeInit~\cite{wu2024freeinit} and FreqPrior~\cite{yuan2025freqprior}, and (ii) search-based methods, including VideoReward~\cite{liu2025improving} and EvoSearch~\cite{he2025scaling}. All methods use a random seed of 42.
\vspace{-1em}
\begin{itemize}
\item \textbf{FreeInit}\cite{wu2024freeinit}: We follow the original setup, using 4 extra sampling iterations and applying a Butterworth filter with a normalised spatial-temporal cutoff frequency of 0.25 as the low-pass filter.
\item \textbf{FreqPrior}\cite{yuan2025freqprior}: Following the paper’s setting, we use 2 extra sampling iterations and the same Butterworth filter configuration as FreeInit. The timestep $t$ is set to 768, and the ratio $cos \theta$ is set to 0.8.
\item \textbf{VideoReward}~\cite{liu2025improving}: We adopt a best-of-N search strategy. Specifically, we sample 4 different initial noise tensors, denoise each through the full trajectory, and decode them into videos. The video reward model is then used to evaluate video quality, motion quality, and text alignment, after which we select the highest-scoring video as the final output.
\item \textbf{EvoSearch}~\cite{he2025scaling}: We employ the configuration with the best performance, setting the population size schedule to \{6, 3, 3\} and the evolution schedule to \{5, 20\}. Following the paper, we use the DPM++ solver, with an elite size of 3 and a mutation rate of 0.2.
\end{itemize}

\vspace{-1.5em}
\section{Additional Experimental Results}
\vspace{-0.5em}

In this section, we provide additional experimental results to further validate the latent reward model and the search procedure of our method.

\vspace{-1em}
\subsection{Latent Reward Model Analysis and Ablations}
\vspace{-0.5em}

\begin{table}[!t]
\begin{center}
\vspace{-1.0em}
\caption{Comparison of VQ, MQ, and TA accuracy across different credit assignment strategies and denoising steps.}
\vspace{-0.5em}
\label{tab_c1}
\resizebox{\columnwidth}{!}{%
\begin{tabular}{@{}c|cccc|cccc|cccc|>{\columncolor{gray!15}}c >{\columncolor{gray!15}}c >{\columncolor{gray!15}}c >{\columncolor{gray!15}}c}
\toprule
\multirow{2}{*}{\begin{tabular}[c]{@{}c@{}} Steps\end{tabular}} 
   & \multicolumn{4}{c|}{VQ Accuracy} 
   & \multicolumn{4}{c|}{MQ Accuracy} 
   & \multicolumn{4}{c|}{TA Accuracy} 
   & \multicolumn{4}{c}{\cellcolor{gray!25}Average Accuracy}\\ 
\cmidrule(l){2-17} 
   & Uni.& Exp.& L2 & Cos. & Uni.& Exp.& L2 & Cos. & Uni.& Exp.& L2 & Cos. & Uni.& Exp.& L2 & Cos.  \\
\midrule
10 & 72.48 & 72.72 & 74.81 & 73.57 & 67.36 & 68.27 & 68.68 & 70.99 & 81.15 & 80.57 & 80.56 & 80.54 & 73.66 & 73.85 & 74.68 & 75.03 \\
15 & 73.56 & 73.94 & 75.20 & 74.76 & 69.40 & 69.84 & 70.34 & 71.22 & 81.26 & 80.66 & 81.22 & 80.85 & 74.74 & 74.81 & 75.59 & 75.61 \\
20 & 73.85 & 73.13 & 75.56 & 75.97 & 69.38 & 69.15 & 70.69 & 72.11 & 81.46 & 81.25 & 81.13 & 81.10 & 74.90 & 74.51 & 75.79 & 76.39 \\
25 & 73.88 & 73.42 & 76.07 & 76.96 & 68.86 & 69.36 & 70.74 & 72.67 & 81.36 & 81.19 & 81.47 & 81.31 & 74.70 & 74.66 & 76.09 & 76.98 \\
30 & 74.27 & 73.77 & 76.76 & 77.10 & 68.63 & 69.82 & 70.72 & 73.07 & 81.27 & 81.34 & 81.58 & 81.07 & 74.72 & 74.98 & 76.35 & 77.08 \\ 
\midrule
Average & 73.61 & 73.40 & \textbf{75.68} & 75.67 & 68.73 & 69.29 & 70.23 & \textbf{72.01} & 81.30 & 81.00 & \textbf{81.19} & 80.97 & 74.54 & 74.56 & 75.70 & \textbf{76.22} \\
\bottomrule
\end{tabular}
}
\vspace{-2em}
\end{center}
\end{table}
\begin{table}[!t]
\begin{center}
\caption{VBench-2.0 evaluation results per dimension under different credit assignment strategies.}
\label{tab_c2}
\vspace{-2em}
\resizebox{\columnwidth}{!}{%
\begin{tabular}{c|cccccc}
\toprule
Methods & Diversity & Composition & \begin{tabular}[c]{@{}c@{}}Motion\\ Rationality\end{tabular} & \begin{tabular}[c]{@{}c@{}}Instance\\ Preservation\end{tabular} & \begin{tabular}[c]{@{}c@{}}Dynamic\\ Spatial\\ Relationship\end{tabular} & \begin{tabular}[c]{@{}c@{}}Dynamic\\ Attribute\end{tabular} \\ 
\midrule
Baseline          & 65.85 & 41.76 & 25.29 & 85.96 & 30.92 & 11.36 \\ 
\midrule
Uniform           & 68.49 & 41.46 & 27.01 & 77.78 & 30.72 & 11.72 \\
Exponential       & 67.68 & 41.31 & 28.74 & 75.44 & 29.95 & 12.45 \\
L2 Error          & 66.65 & 44.90 & 29.59 & 83.68 & 29.57 & 15.38 \\
Cosine Similarity & 68.33 & 47.90 & 32.76 & 85.97 & 29.95 & 15.38 \\
\midrule 
\midrule
Methods & \begin{tabular}[c]{@{}c@{}}Motion\\ Order\\ Understanding\end{tabular} & \begin{tabular}[c]{@{}c@{}}Human\\ Interaction\end{tabular} & \begin{tabular}[c]{@{}c@{}}Complex\\ Landscape\end{tabular} & \begin{tabular}[c]{@{}c@{}}Complex\\ Plot\end{tabular} & \begin{tabular}[c]{@{}c@{}}Camera\\ Motion\end{tabular} & \begin{tabular}[c]{@{}c@{}}Human\\ Anatomy\end{tabular} \\ 
\midrule
Baseline          & 16.84 & 54.00 & 16.22 & 11.33 & 13.27 & 80.67 \\ 
\midrule
Uniform           & 19.53 & 46.00 & 17.33 & 10.89 & 13.89 & 79.93 \\
Exponential       & 19.53 & 48.33 & 18.44 & 10.03 & 12.65 & 78.86 \\
L2 Error          & 19.19 & 50.11 & 18.89 & 9.90  & 12.35 & 76.24 \\
Cosine Similarity & 20.88 & 50.67 & 17.56 & 11.44 & 12.96 & 77.97 \\
\midrule
\midrule
Methods & \begin{tabular}[c]{@{}c@{}}Human\\ Identity\end{tabular} & \begin{tabular}[c]{@{}c@{}}Human\\ Clothes\end{tabular} & Mechanics & Thermotics & Material & \begin{tabular}[c]{@{}c@{}}Multi-View\\ Consistency\end{tabular} \\ 
\midrule
Baseline          & 68.41 & 97.24 & 60.33 & 55.47 & 32.84 & 35.28 \\ 
\midrule
Uniform           & 77.73 & 97.27 & 55.91 & 58.33 & 43.33 & 33.43 \\
Exponential       & 74.79 & 95.45 & 56.91 & 56.55 & 43.04 & 40.68 \\
L2 Error          & 75.09 & 95.02 & 64.52 & 54.07 & 36.49 & 32.71 \\
Cosine Similarity & 74.37 & 95.43 & 62.50 & 50.37 & 37.97 & 34.93 \\
\bottomrule
\end{tabular}
}
\vspace{-2.5em}
\end{center}
\end{table}
\begin{figure}[!t]
\centering
    \includegraphics[width=0.85\textwidth]{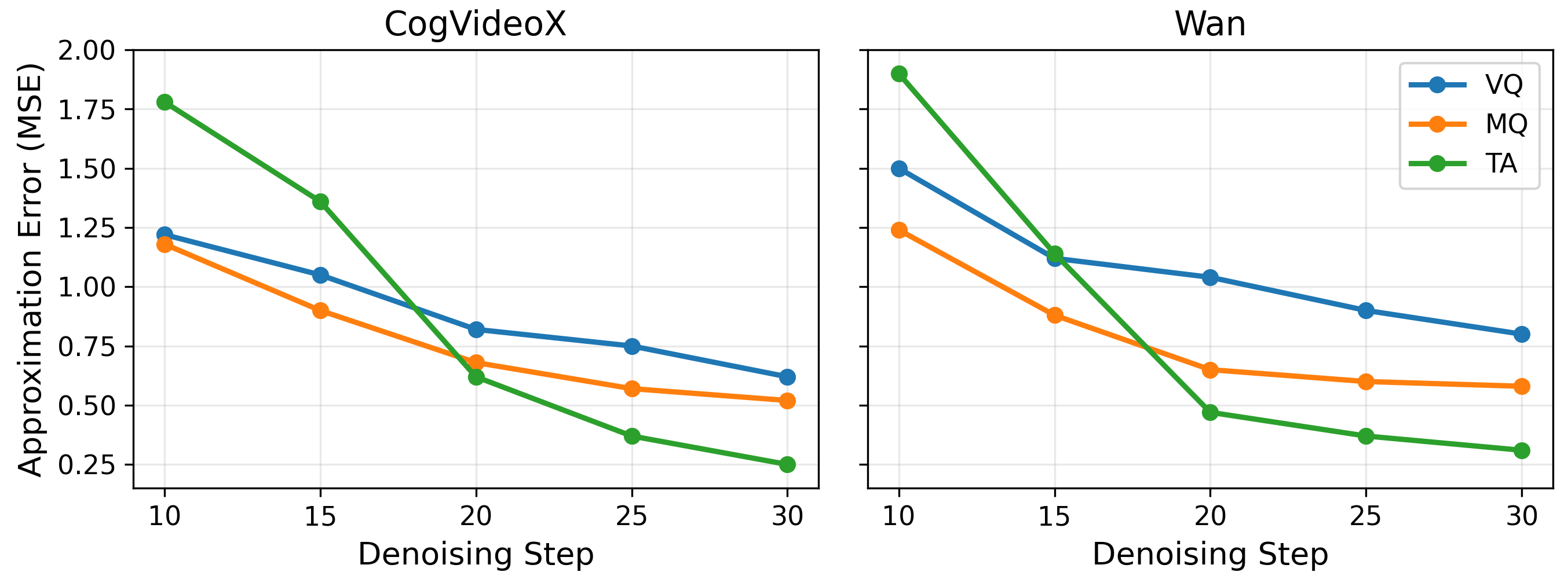}
    \vspace{-1em}
    \caption{Empirical approximation error of cosine-similarity credit assignment on the test set. We convert predicted latent rewards back to final video-level reward estimates and compute MSE against the true final video rewards. The error decreases along denoising on both Wan2.1-1.3B and CogVideoX-2B, indicating that later latents are more predictive of final rewards.}
    \label{approx_error}
    \vspace{-2.5em}
\end{figure}

\noindent \textbf{Credit Assignment Strategies.} Tables~\ref{tab_c1} and~\ref{tab_c2} evaluate different latent credit-assignment strategies from both reward prediction accuracy and downstream video quality perspectives. Uniform weighting consistently underperforms across denoising steps, indicating that treating all latent states equally fails to capture their varying semantic relevance during the generation process. Exponential weighting improves performance at earlier steps but exhibits less stability across later stages, suggesting that aggressively prioritizing early latents may introduce bias. In contrast, L2-error and cosine-similarity based strategies achieve consistently stronger performance across VQ, MQ, and TA accuracy. Notably, cosine similarity yields the best overall balance, achieving the highest average accuracy across denoising steps and translating into stronger VBench-2.0 performance in motion rationality, instance preservation, and motion order understanding. This indicates that similarity-based assignment provides a more reliable mechanism for identifying semantically meaningful latent states and enables more stable supervision for the latent reward model.

\noindent \textbf{Empirical Approximation Error of Cosine-Similarity Credit Assignment.}~We further analyse the approximation error introduced by using cosine similarity to assign video-level rewards to intermediate latents. Specifically, for each intermediate latent in the test set, we predict latent rewards with the latent reward model and convert them back to final video-level reward estimates according to the cosine-similarity credit-assignment rule. We then compute the MSE between the recovered reward estimates and the true final video rewards.

As shown in Fig.~\ref{approx_error}, the approximation error consistently decreases along the denoising trajectory on both Wan2.1-1.3B and CogVideoX-2B. This indicates that later latents become increasingly predictive of the final visual quality, motion quality, and text-alignment rewards. The same trend holds across different backbones, suggesting that cosine-similarity credit assignment provides a meaningful proxy for supervising intermediate latent states.

\begin{figure}[!t]
    \centering 
    \includegraphics[width=0.97\textwidth]{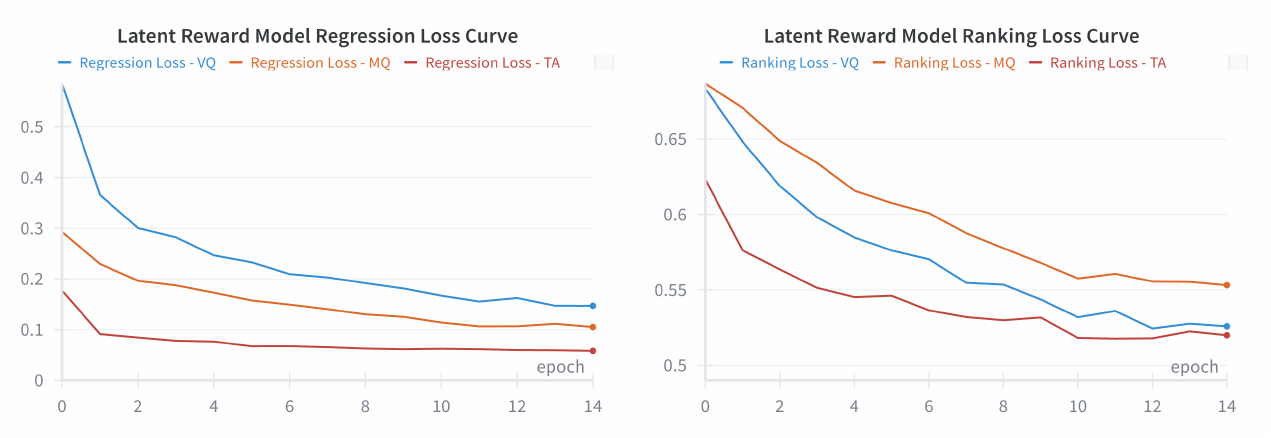}
    \vspace{-1em}
    \caption{Training curves of the latent reward model: regression loss and reference loss.} 
    \label{fig_c1}
    \vspace{-1em}
\end{figure}
\begin{table}[!t]
\begin{center}
\caption{Comparison of VQ, MQ, and TA accuracy across different loss function settings and denoising steps. PL denotes preference loss.}
\vspace{-1em}
\label{tab_c3}
\resizebox{0.8\columnwidth}{!}{%
\begin{tabular}{@{}c|cc|cc|cc|>{\columncolor{gray!15}}c >{\columncolor{gray!15}}c}
\toprule
\multirow{2}{*}{\begin{tabular}[c]{@{}c@{}}Latent\\ Denoising Steps\end{tabular}} 
& \multicolumn{2}{c|}{VQ Accuracy} 
& \multicolumn{2}{c|}{MQ Accuracy} 
& \multicolumn{2}{c|}{TA Accuracy} 
& \multicolumn{2}{c}{\cellcolor{gray!25}Average Accuracy}\\ 
\cmidrule(l){2-9} 
& w/o PL & w/ PL & w/o PL & w/ PL & w/o PL & w/ PL & \cellcolor{gray!15}w/o PL & \cellcolor{gray!15}w/ PL \\
\midrule
10 & 71.50 & \textbf{73.57} & 66.46 & \textbf{70.99} & 77.84 & \textbf{80.54} & 71.93 & \textbf{75.03} \\
15 & 72.51 & \textbf{74.76} & 67.88 & \textbf{71.22} & 78.50 & \textbf{80.85} & 72.96 & \textbf{75.61} \\
20 & 72.99 & \textbf{75.97} & 69.20 & \textbf{72.11} & 78.10 & \textbf{81.10} & 73.43 & \textbf{76.39} \\
25 & 73.52 & \textbf{76.96} & 68.63 & \textbf{72.67} & 78.18 & \textbf{81.31} & 73.44 & \textbf{76.98} \\
30 & 74.49 & \textbf{77.10} & 69.49 & \textbf{73.07} & 77.62 & \textbf{81.07} & 73.87 & \textbf{77.08} \\ 
\bottomrule
\end{tabular}
}
\end{center}
\vspace{-3em}
\end{table}

\begin{table}[!t]
\begin{center}
\caption{Impact of search budget, temperature, and search schedule on VBench-2.0 performance across all dimensions.}
\vspace{-2em}
\label{tab_c4}
\resizebox{\columnwidth}{!}{%
\begin{tabular}{c|cccccc}
\toprule
Methods & Diversity & Composition & \begin{tabular}[c]{@{}c@{}}Motion\\ Rationality\end{tabular} & \begin{tabular}[c]{@{}c@{}}Instance\\ Preservation\end{tabular} & \begin{tabular}[c]{@{}c@{}}Dynamic\\ Spatial\\ Relationship\end{tabular} & \begin{tabular}[c]{@{}c@{}}Dynamic\\ Attribute\end{tabular} \\ 
\midrule
Baseline & 65.85 & 41.76 & 25.29 & 85.96 & 30.92 & 11.36 \\ 
\midrule
Search Budget $N=4$	 & 70.32 & 45.08 & 31.03 & 78.36 & 31.41 & 15.02 \\
Search Budget $N=6$	 & 68.33 & 47.90 & 32.76 & 85.97 & 29.95 & 15.38 \\
Search Budget $N=8$	 & 69.19 & 47.75 & 31.61 & 86.12 & 29.47 & 12.46 \\
\midrule
Temperature $\tau=0.5$ & 66.99 & 46.49 & 29.89 & 86.61 & 28.50 & 13.92 \\
Temperature $\tau=1.0$ & 68.33 & 47.90 & 32.76 & 85.97 & 29.95 & 15.38 \\
Temperature $\tau=2.0$ & 69.19 & 47.64 & 31.61 & 84.85 & 24.15 & 15.38 \\ 
\midrule
\{10, 15\}              & 69.07 & 46.94 & 31.61 & 80.70 & 30.43 & 15.75 \\
\{10, 15, 20\}          & 68.33 & 47.90 & 32.76 & 85.97 & 29.95 & 15.38 \\
\{10, 15, 20, 25\}      & 68.13 & 46.46 & 28.16 & 85.44 & 32.85 & 17.95 \\
\{10, 15, 20, 25, 30\}  & 65.23 & 45.57 & 29.31 & 85.95 & 29.47 & 15.38 \\
\midrule 
\midrule
Methods & \begin{tabular}[c]{@{}c@{}}Motion\\ Order\\ Understanding\end{tabular} & \begin{tabular}[c]{@{}c@{}}Human\\ Interaction\end{tabular} & \begin{tabular}[c]{@{}c@{}}Complex\\ Landscape\end{tabular} & \begin{tabular}[c]{@{}c@{}}Complex\\ Plot\end{tabular} & \begin{tabular}[c]{@{}c@{}}Camera\\ Motion\end{tabular} & \begin{tabular}[c]{@{}c@{}}Human\\ Anatomy\end{tabular} \\ \midrule
Baseline & 16.84 & 54.00 & 16.22 & 11.33 & 13.27 & 80.67 \\ 
\midrule
Search Budget $N=4$	 & 19.87 & 48.33 & 18.00 & 10.23 & 11.11 & 78.42 \\
Search Budget $N=6$	 & 20.88 & 50.67 & 17.56 & 11.44 & 12.96 & 77.97 \\
Search Budget $N=8$	 & 20.51 & 50.67 & 18.21 & 9.65 & 14.82 & 78.51 \\
\midrule
Temperature $\tau=0.5$ & 18.86 & 51.67 & 17.11 & 9.14 & 13.58 & 78.18 \\
Temperature $\tau=1.0$ & 20.88 & 50.67 & 17.56 & 11.44 & 12.96 & 77.97 \\
Temperature $\tau=2.0$ & 17.51 & 51.67 & 17.56 & 10.12 & 14.51 & 79.31 \\ 
\midrule
\{10, 15\}              & 17.85 & 49.00 & 17.11 & 9.01 & 15.43 & 78.92 \\
\{10, 15, 20\}          & 20.88 & 50.67 & 17.56 & 11.44 & 12.96 & 77.97 \\
\{10, 15, 20, 25\}      & 14.82 & 50.33 & 17.33 & 10.54 & 14.81 & 78.66 \\
\{10, 15, 20, 25, 30\}  & 15.82 & 49.00 & 16.22 & 10.44 & 14.81 & 77.73 \\
\midrule
\midrule
Methods & \begin{tabular}[c]{@{}c@{}}Human\\ Identity\end{tabular} & \begin{tabular}[c]{@{}c@{}}Human\\ Clothes\end{tabular} & Mechanics & Thermotics & Material & \begin{tabular}[c]{@{}c@{}}Multi-View\\ Consistency\end{tabular} \\ 
\midrule
Baseline & 68.41 & 97.24 & 60.33 & 55.47 & 32.84 & 35.28 \\ 
\midrule
Search Budget $N=4$	 & 76.51 & 95.95 & 56.57 & 52.24 & 41.56 & 33.76 \\
Search Budget $N=6$	 & 74.37 & 95.43 & 62.50 & 50.37 & 37.97 & 34.93 \\
Search Budget $N=8$	 & 77.10 & 96.38 & 62.11 & 56.55 & 35.23 & 34.29 \\
\midrule
Temperature $\tau=0.5$ & 79.04 & 96.38 & 56.67 & 51.49 & 38.96 & 38.64 \\
Temperature $\tau=1.0$ & 74.37 & 95.43 & 62.50 & 50.37 & 37.97 & 34.93 \\
Temperature $\tau=2.0$ & 72.95 & 94.09 & 58.54 & 51.13 & 36.14 & 39.46 \\ 
\midrule
$\{10, 15\}$            & 73.17 & 95.05 & 59.50 & 49.21 & 38.96 & 21.20 \\
$\{10, 15, 20\}$        & 74.37 & 95.43 & 62.50 & 50.37 & 37.97 & 34.93 \\
$\{10, 15, 20, 25\}$    & 72.61 & 96.86 & 60.80 & 52.59 & 35.44 & 37.61 \\
$\{10, 15, 20, 25, 30\}$ & 73.33 & 94.09 & 61.75 & 50.56 & 36.71 & 32.34 \\
\bottomrule
\end{tabular}
}
\end{center}
\vspace{-3em}
\end{table}

\noindent \textbf{Training Curves of the Latent Reward Model.} Figure \ref{fig_c1} shows the training dynamics of the latent reward model under regression and ranking objectives. Both losses exhibit stable convergence, with ranking loss decreasing steadily alongside regression error. The smooth optimization behavior indicates that the joint training objective does not introduce instability and effectively enhances the model’s discriminative capability without sacrificing regression fidelity.

\noindent \textbf{Effect of Preference Loss.} Table \ref{tab_c3} demonstrates that incorporating preference loss consistently improves reward prediction accuracy across all denoising steps. Compared to the regression-only objective, preference supervision leads to noticeable gains in MQ and TA accuracy, resulting in an overall average improvement of approximately 1–2\%. The advantage becomes more evident at later denoising stages, suggesting that preference learning helps refine fine-grained distinctions between latent trajectories when the generation process approaches semantic convergence.

\vspace{-1.5em}
\subsection{Sensitivity to Search Hyperparameters}
\vspace{-0.5em}
We analyse the robustness of LatSearch with respect to three key hyperparameters.

\textit{(i) Search Budget $N$.} Increasing $N$ enhances exploration but raises compute cost. As shown in Table~\ref{tab_c4}, performance improves monotonically from $N{=}4$ to $N{=}8$, confirming that the search mechanism scales reliably with additional candidates.

\textit{(ii) Temperature $\tau$.} Varying $\tau \in \{0.5, 1.0, 2.0\}$ leads to only small changes in VBench-2.0 scores (Table~\ref{tab_c5}), indicating that the resampling step is stable across a wide temperature range.

\textit{(iii) Scoring-timestep schedule $S$.} The choice of scoring timesteps is crucial because reward quality varies substantially across the diffusion trajectory. We avoid applying scoring at very early steps because latents in these stages are dominated by noise and contain almost no semantic content, making reward estimation unstable and non-informative.

Conversely, we do not apply scoring at very late timesteps primarily due to computational cost. For instance, with $N{=}6$ candidates and a DiT forward time of 1.35\,s per step, scoring within the [10,20] interval increases total inference time by approximately 67.5–135\,s (best–worst case). Scoring within [30,40], however, increases the cost to 202–270\,s, resulting in a 2--3 times runtime increase. This conflicts with our motivation of enabling early, efficient search.

For this reason, we intentionally apply search as early as possible once semantically meaningful structure emerges. Our ablations in Table~\ref{tab_c6} confirm this design: too few scoring points (e.g., $\{10,15\}$) provide insufficient temporal coverage, while too many (e.g., $\{10,15,20,25,30\}$) accumulate uncertainty from similarity-derived targets and degrade performance. A moderate, well-spaced mid-range schedule, $\{10,15,20\}$, achieves the best trade-off between discriminative power and computational efficiency.

\begin{table}[!t]
\centering
\caption{Sensitivity analysis of the temperature parameter $\tau$ on VBench-2.0.}
\vspace{-1em}
\label{tab_c5}
\resizebox{0.9\columnwidth}{!}{%
\begin{tabular}{c|c|c|c|c|c|c}
\toprule
Temperature &
Creativity &
Commonsense &
Controllability &
Human Fidelity &
Physics &
Average \\
\midrule
0.5 & 56.74 & 58.25 & 21.83 & 84.53 & 46.44 & \textbf{53.56} \\
1.0 & 58.12 & 59.37 & 22.69 & 82.59 & 46.44 & \textbf{53.84} \\
2.0 & 58.42 & 58.23 & 21.56 & 82.12 & 46.32 & \textbf{53.33} \\
\bottomrule
\end{tabular}
}
\end{table}
\begin{table}[!t]
\centering
\caption{Effect of scoring timestep schedules on VBench-2.0.}
\vspace{-1em}
\label{tab_c6}
\resizebox{\columnwidth}{!}{%
\begin{tabular}{c|c|c|c|c|c|c|c}
\toprule
\begin{tabular}[c]{@{}c@{}}Search \\ Scheduler\end{tabular} &
\begin{tabular}[c]{@{}c@{}}Creativity\end{tabular} &
\begin{tabular}[c]{@{}c@{}}Commonsense\end{tabular} &
\begin{tabular}[c]{@{}c@{}}Controllability\end{tabular} &
\begin{tabular}[c]{@{}c@{}}Human\\ Fidelity\end{tabular} &
\begin{tabular}[c]{@{}c@{}}Physics\end{tabular} &
\begin{tabular}[c]{@{}c@{}}Average\end{tabular} &
\begin{tabular}[c]{@{}c@{}}Inference\\ Time (s)\end{tabular} \\
\midrule
Baseline & 53.81 & 55.63 & 21.99 & 82.11 & 45.98 & \textbf{51.90} & $77.21 \pm 0.26$ \\
\midrule
$\{10, 15\}$ & 58.01 & 56.16 & 22.08 & 82.38 & 42.22 & \textbf{52.17 (+0.27)} & $154.63 \pm 3.06$ \\
$\{10, 15, 20\}$ & 58.12 & 59.37 & 22.69 & 82.59 & 46.44 & \textbf{53.84 (+1.94)} & $182.43 \pm 6.53$ \\
$\{10, 15, 20, 25\}$ & 57.30 & 56.80 & 22.66 & 82.71 & 46.61 & \textbf{53.22 (+1.32)} & $186.47 \pm 8.58$ \\
$\{10, 15, 20, 25, 30\}$ & 55.40 & 57.63 & 21.59 & 81.72 & 45.34 & \textbf{52.34 (+0.44)} & $198.48 \pm 10.99$ \\
\bottomrule
\end{tabular}
}
\end{table}
\begin{table}[!t]
\centering
\caption{Human pairwise preference rates.}
\vspace{-1em}
\label{tab_c7}
\resizebox{0.6\columnwidth}{!}{%
\begin{tabular}{c|c}
\toprule
\textbf{Criterion} & \textbf{Preference for LatSearch} \\
\midrule
Visual Quality   & 72.15\% \\
Motion Quality   & 75.29\% \\
Video-Text Alignment   & 78.51\% \\
\midrule
\textbf{Average} & \textbf{75.32\%} \\
\bottomrule
\end{tabular}
}
\vspace{-1.5em}
\end{table}

\vspace{-1em}
\subsection{Human Preference Study}
To further validate the perceptual quality of LatSearch beyond VBench-2.0 metrics, we conducted a pairwise human preference study. Specifically, we sampled 40 video pairs generated by the baseline model and LatSearch across diverse categories, including animals, humans, natural scenery, and architecture. Each pair was evaluated by 30 participants, who were asked to choose which video was better along three key criteria:

\begin{itemize}
\item \textit{Visual Quality:} clarity, level of detail, and absence of artifacts;
\item \textit{Motion Quality:} temporal consistency, smoothness, and physical realism;
\item \textit{Video–Text Alignment:} fidelity of subjects, actions, and scenes to the prompt.
\end{itemize}

Results in Table~\ref{tab_c7} indicate that participants expressed a clear preference for LatSearch across all criteria, demonstrating that the improvements observed in automated metrics also align with human subjective judgment.

\vspace{-0.5em}
\subsection{Full VBench-2.0 Results Across All Methods and Backbones}
\vspace{-0.5em}
For completeness, we present the per-dimension VBench-2.0 results across all methods and backbones in Table \ref{tab_c8}, complementing the overall scores reported in the main paper.
\begin{table}[!t]
\begin{center}
\caption{VBench-2.0 evaluation results per dimension across different methods. $^\dagger$ indicates the use of DPM-Solver++.}
\vspace{-1em}
\label{tab_c8}
\resizebox{\columnwidth}{!}{%
\begin{tabular}{l|cccccc}
\toprule
Methods & Diversity & Composition & \begin{tabular}[c]{@{}c@{}}Motion\\ Rationality\end{tabular} & \begin{tabular}[c]{@{}c@{}}Instance\\ Preservation\end{tabular} & \begin{tabular}[c]{@{}c@{}}Dynamic\\ Spatial\\ Relationship\end{tabular} & \begin{tabular}[c]{@{}c@{}}Dynamic\\ Attribute\end{tabular} \\ 
\midrule
Baseline (Wan2.1-1.3B) & 65.85 & 41.76 & 25.29 & 85.96 & 30.92 & 11.36 \\ 
\midrule
+ FreeInit & 55.49 & 38.11 & 27.01 & \textbf{91.81} & 29.95 & 9.16 \\
+ FreqPrior & 66.35 & 41.04 & 27.01 & 84.21 & 28.50 & 6.59 \\
+ VideoReward & 62.74 & 47.40 & 31.61 & 84.21 & 31.88 & 13.92 \\
+ EvoSearch$^\dagger$ & \textbf{71.88} & 46.62 & \textbf{35.05} & 87.13 & \textbf{35.75} & 11.72 \\ 
\midrule
+ LatSearch (w/o RGRP) & 66.18 & 46.70 & 29.89 & 87.13 & 33.82 & 13.92 \\
+ LatSearch & 68.33 & \textbf{47.90} & 32.76 & 85.97 & 29.95 & \textbf{15.38} \\
+ LatSearch$^\dagger$ & 69.19 & 45.87 & 28.16 & 86.55 & 29.95 & 13.55 \\ 
\midrule
Baseline (Wan2.1-14B) & 62.34 & \textbf{48.08} & 28.73 & 83.62 & 24.15 & \textbf{13.55} \\ 
+ LatSearch  & \textbf{66.18} & 46.39 & \textbf{29.31} & \textbf{85.97} & \textbf{25.60} & 13.19 \\
\midrule 
\midrule
Methods & \begin{tabular}[c]{@{}c@{}}Motion\\ Order\\ Understanding\end{tabular} & \begin{tabular}[c]{@{}c@{}}Human\\ Interaction\end{tabular} & \begin{tabular}[c]{@{}c@{}}Complex\\ Landscape\end{tabular} & \begin{tabular}[c]{@{}c@{}}Complex\\ Plot\end{tabular} & \begin{tabular}[c]{@{}c@{}}Camera\\ Motion\end{tabular} & \begin{tabular}[c]{@{}c@{}}Human\\ Anatomy\end{tabular} \\ \midrule
Baseline & 16.84 & 54.00 & 16.22 & 11.33 & 13.27 & 80.67 \\ 
\midrule
+ FreeInit & 17.17 & 53.00 & 18.00 & \textbf{11.85} & 15.12 & 81.53 \\
+ FreqPrior & 16.50 & 48.33 & \textbf{21.33} & 11.63 & 9.57 & 79.61 \\
+ VideoReward & 17.51 & 54.33 & 18.22 & 11.69 & 14.51 & 78.93 \\
+ EvoSearch$^\dagger$ & \textbf{21.88} & \textbf{65.67} & 19.78 & 9.94 & \textbf{18.52} & \textbf{82.24} \\ 
\midrule
+ LatSearch (w/o RGRP) & 18.86 & 48.67 & 17.78 & 9.31 & 13.58 & 79.54 \\
+ LatSearch & 20.88 & 50.67 & 17.56 & 11.44 & 12.96 & 77.97 \\
+ LatSearch$^\dagger$ & 19.87 & 57.33 & 17.78 & 10.69 & \textbf{18.52} & 79.62 \\ 
\midrule
Baseline (Wan2.1-14B) & \textbf{22.29} & 55.33 & 14.44 & \textbf{12.90} & 9.87 & 87.24 \\ 
+ LatSearch  & 18.58 & \textbf{61.67} & \textbf{16.22} & 12.17 & \textbf{10.19} & \textbf{89.53} \\
\midrule
\midrule
Methods & \begin{tabular}[c]{@{}c@{}}Human\\ Identity\end{tabular} & \begin{tabular}[c]{@{}c@{}}Human\\ Clothes\end{tabular} & Mechanics & Thermotics & Material & \begin{tabular}[c]{@{}c@{}}Multi-View\\ Consistency\end{tabular} \\ 
\midrule
Baseline & 68.41 & 97.24 & 60.33 & 55.47 & 32.84 & 35.28 \\ 
\midrule
+ FreeInit & 75.22 & \textbf{98.93} & 52.14 & 44.48 & 32.47 & 13.52 \\
+ FreqPrior & 73.55 & 97.69 & 53.09 & 53.03 & 27.27 & 21.04 \\
+ VideoReward & 71.33 & 96.39 & 60.17 & \textbf{57.97} & 33.33 & 31.24 \\
+ EvoSearch$^\dagger$ & \textbf{80.19} & 97.77 & 54.76 & 47.65 & 35.16 & 29.63 \\ 
\midrule
+ LatSearch (w/o RGRP)& 76.15 & 97.30 & 53.97 & 51.52 & \textbf{40.26} & 36.39 \\
+ LatSearch  & 74.37 & 95.43 & \textbf{62.50} & 50.37 & 37.97 & 34.93 \\
+ LatSearch$^\dagger$ & 74.22 & 98.18 & 61.72 & 46.57 & 37.50 & \textbf{67.85} \\ 
\midrule
Baseline (Wan2.1-14B) & 83.82 & 98.16 & \textbf{52.17} & \textbf{47.88} & 36.20 & \textbf{23.60} \\ 
+ LatSearch  & \textbf{85.28} & \textbf{99.54} & 52.03 & 47.14 & \textbf{38.81} & 22.71 \\
\bottomrule
\end{tabular}
}
\vspace{-2em}
\end{center}
\end{table}

\section{Qualitative Comparisons}
\begin{figure}[!t]
    \vspace{-1em}
    \centering 
    \includegraphics[width=0.99\textwidth]{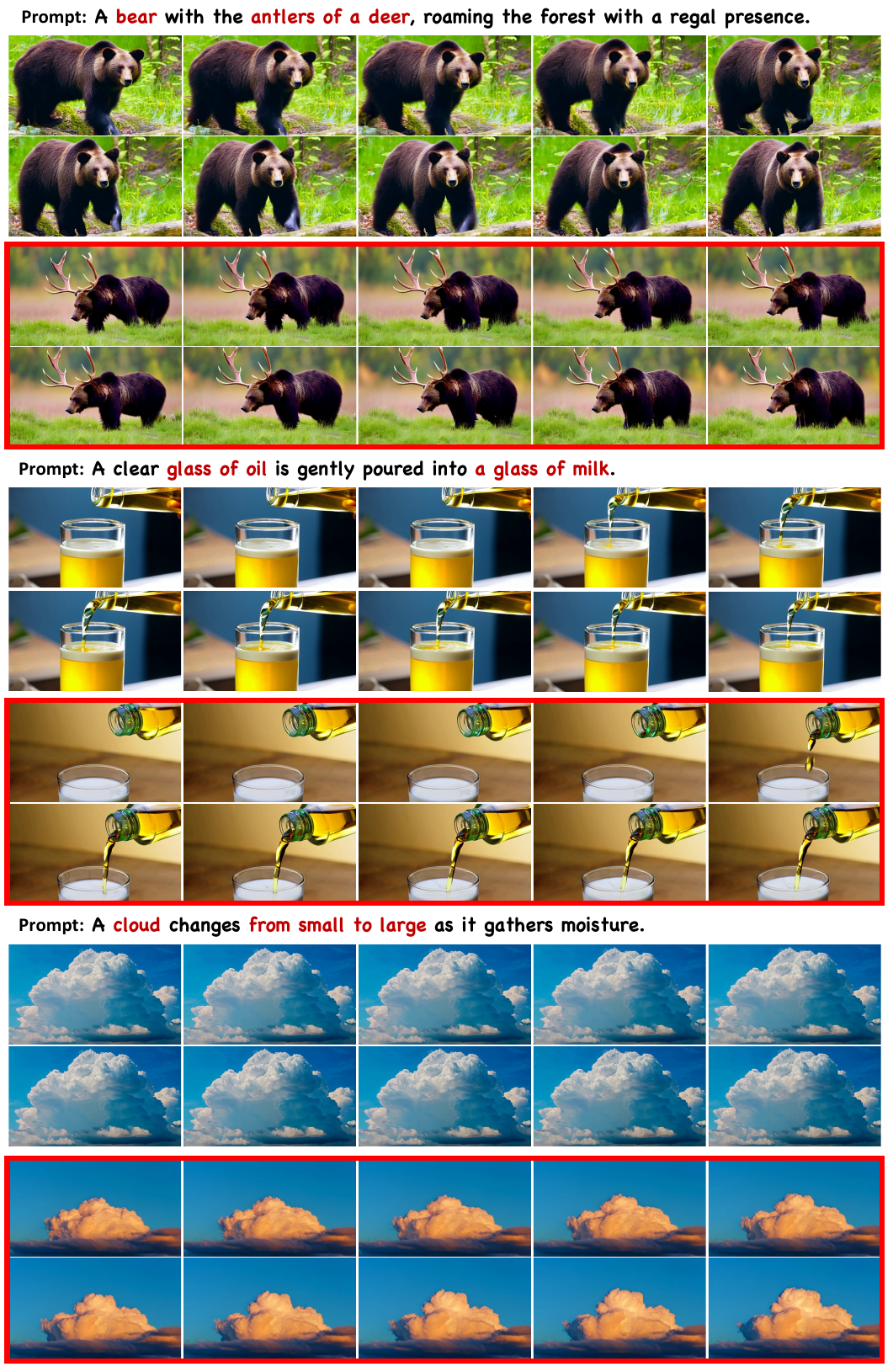}
    \vspace{-1em}
    \caption{Comparison of text-to-video generation results between the baseline model (top) and LatSearch (bottom) for each prompt.}
    \label{fig4}
\end{figure}
\begin{figure}[!t]
    \vspace{-1em}
    \centering 
    \includegraphics[width=0.99\textwidth]{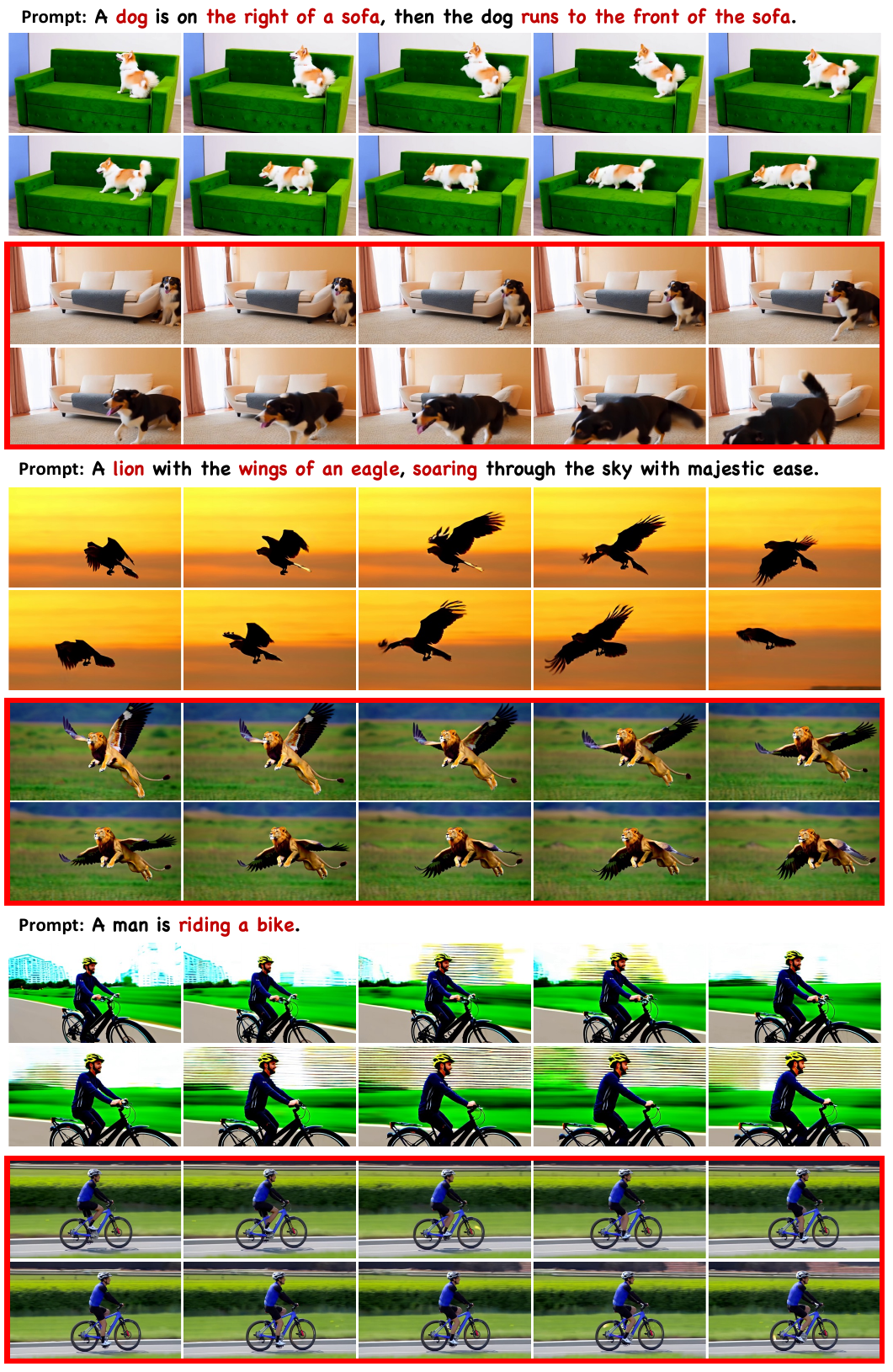}
    \vspace{-1em}
    \caption{Comparison of text-to-video generation results between the baseline model (top) and LatSearch (bottom) for each prompt.}
    \label{fig4}
\end{figure}
\begin{figure}[!t]
    \vspace{-1em}
    \centering 
    \includegraphics[width=0.99\textwidth]{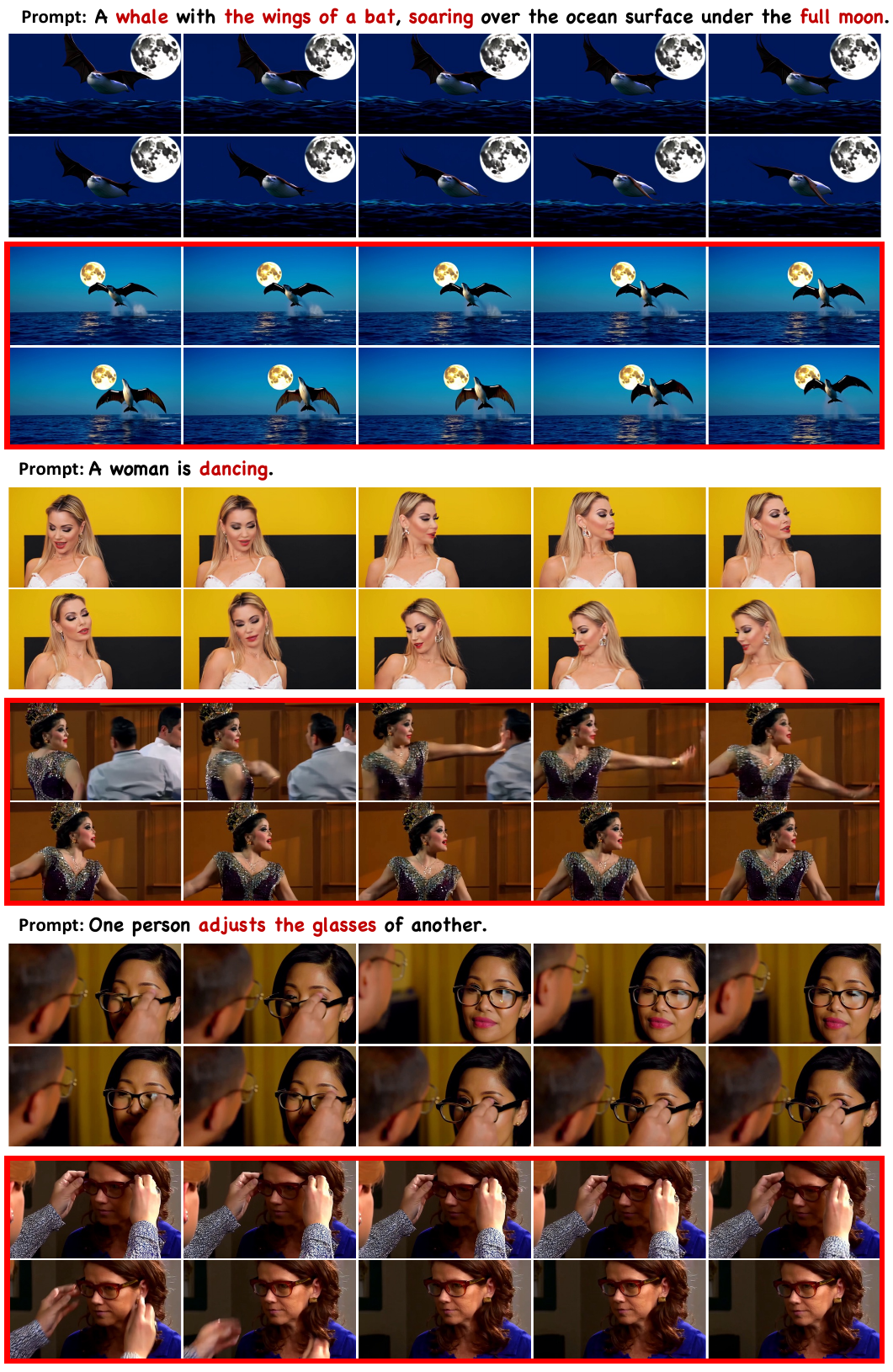}
    \vspace{-1em}
    \caption{Comparison of text-to-video generation results between the baseline model (top) and LatSearch (bottom) for each prompt.}
    \label{fig4}
\end{figure}

\end{document}